\documentclass[10pt]{article} %

\usepackage[accepted]{rlj} %

\usepackage{amssymb}            %
\usepackage{mathtools}          %
\usepackage{mathrsfs}           %
\usepackage{graphicx,multicol}
\usepackage{subcaption}         %
\usepackage[space]{grffile}     %
\usepackage{url}                %
\usepackage{lipsum}             %

\usepackage{float}
\usepackage{wrapfig}
\usepackage{tikz}
\usetikzlibrary{positioning, arrows.meta, fit, backgrounds}

\title{Forager: a lightweight testbed for continual learning with partial observability in reinforcement learning
}

\setrunningtitle{Forager: a lightweight testbed for continual learning with partial observability in RL
}

\author{Steven Tang\textsuperscript{1,2}, Xinze Xiong\textsuperscript{1,2}, Anna Hakhverdyan\textsuperscript{1,2}, Andrew Patterson\textsuperscript{1,2}, \ \ Jacob Adkins\textsuperscript{1,2}, Jiamin He\textsuperscript{1,2}, Esraa Elelimy\textsuperscript{1,2}, Parham Mohammad Panahi\textsuperscript{1,2}, Martha White\textsuperscript{1,2,3}, Adam White\textsuperscript{1,2,3}}

\emails{\{stang5, xinze5, hakhverd, ap3, jadkins, jiamin12, elelimy, parham1, whitem, amw8\}@ualberta.ca}

\affiliations{
$^{1}$\textbf{Department of Computing Science, University of Alberta, Canada}\\
$^{2}$\textbf{Alberta Machine Intelligence Institute (Amii)}\\
$^{3}$\textbf{CIFAR AI Chair}
}

\contribution{
    A new testbed for continual learning research that is resource efficient allowing for fast long-running experiments and rapid experiment iteration. Forager is configurable with user-specified parameters that allow fine control of the degree of partial observability and problem difficulty. We provide several sample use cases, including one that uses random generation to create an unending stream of new tasks to ensure that agents must continue to learn forever and never converge as highlighted as a key feature of continual learning in the literature~\citep{abel2023definition}. This is all wrapped up in an open-source repository.
    }
    {
    Jelly Bean World~\citep{platanios2020jelly} was designed under similar principles, but JBW has technical issues in terms of memory usage demonstrated in the introduction. AgarCL~\citep{mohamed2026cell} is another related benchmark, but successful learning has not been demonstrated in this domain. Forager is not realistic and the action and observation space are simple. Forager has no physics, nor other features of rich game-like environments. This is by design. There are plenty of complex environments available to RL researchers. The goal here is to balance rapid prototyping of ideas with just the right level of complexity to challenge current algorithms. 
    }

\contribution{
    We highlight and demonstrate the role of state construction in continual RL. We show that feed forward agents exhibit loss of plasticity which can be mitigated to some degree by recent algorithms. However, simple memory traces and recurrent PPO agents can both mitigate some partial observability and exhibit continual learning and tracking. This highlights the need for new recurrent CRL algorithms.
    }
    {
    Recent work demonstrated that recurrent PPO achieves reliable foraging in ForageWorld~\citep{simmons2025deep}, an extension of Craftax~\citep{matthews2024craftax}, but this required access to privileged state information and many parallel asynchronous actors. Here we focus on continual learning and state construction in the single stream setting as it remains a key challenge in CRL~\citep{abel2023definition,khetarpal2022towards} and asynchronous training can mask this challenge~\citep{mayor2025the}. 
    }


\summary{
In continual reinforcement learning (CRL), good performance requires never-ending learning, acting, and exploration in a big, partially observable world. Most CRL experiments have focused on loss of plasticity---the inability to keep learning---in one-off experiments where some unobservable non-stationarity is added to classic fully observable MDPs. Further, these experiments rarely consider the role of partial observability and the importance of CRL agents that use memory or recurrence. One potential reason for this focus on mitigating loss of plasticity without considering partial observability is that many partially-observable CRL environments are prohibitively expensive. In this paper, we introduce Forager, a lightweight partially-observable CRL environment with a constant memory footprint. We provide a set of experiments and sample tasks demonstrating that Forager is challenging for current CRL agents and yet also allows for in-depth study of those agents. We demonstrate that agents exhibit loss of plasticity, proposed mitigations can help, but that most useful is to leverage state construction. We conclude with a variant of Forager that generates an unending  stream of new tasks to learn that clearly highlights the limitations of current CRL agents.
}

\begin{document}

\makeCover  %
\maketitle  %

\begin{abstract}
In continual reinforcement learning (CRL), good performance requires never-ending learning, acting, and exploration in a big, partially observable world. Most CRL experiments have focused on loss of plasticity---the inability to keep learning---in one-off experiments where some unobservable non-stationarity is added to classic fully observable MDPs. Further, these experiments rarely consider the role of partial observability and the importance of CRL agents that use memory or recurrence. One potential reason for this focus on mitigating loss of plasticity without considering partial observability is that many partially-observable CRL environments are prohibitively expensive. In this paper, we introduce Forager, a lightweight partially-observable CRL environment with a constant memory footprint. We provide a set of experiments and sample tasks demonstrating that Forager is challenging for current CRL agents and yet also allows for in-depth study of those agents. We demonstrate that agents exhibit loss of plasticity, proposed mitigations can help, but that most useful is to leverage state construction. We conclude with a variant of Forager that generates an unending  stream of new tasks to learn that clearly highlights the limitations of current CRL agents. %
\end{abstract}

\section{Introduction}
\label{sec:intro} 

Continual Reinforcement Learning (CRL) is the study of agents that must continue to learn forever in order to perform well. 
In the real world, a person or an agent has a limited way in which it observes the world: it can only see, hear, and touch things in its immediate vicinity---the world is partially observable. In addition, there are effectively an infinite number of things to learn about. 
This view-point, called the \emph{Big World Hypothesis}~\citep{javed2024big}, is often summarized as the agent being much smaller than the environment. 
In such situations, the agent cannot represent the optimal policy and any world model will be approximate. The world is partially observable and from the agent's perspective often appears non-stationary. The best approach is to construct an internal state to mitigate partial observability (e.g., recurrent architectures) and to track~\citep{sutton2007role} to learn continually in deployment~\citep{abbas2023loss,dohare2024loss}.

Most work in reinforcement learning (RL) focuses on fully observable benchmarks where the agent is much larger than the environment. For example, the Arcade Learning Environment (ALE) \citep{bellemare2013arcade} requires only 128 bytes of RAM and 4KB of ROM, whereas Nature DQN (a small agent by today's standards) contains over 1.7 million learnable parameters \citep{mnih2015human}. 
In addition, frame-stacking largely mitigates partial observability in ALE, while other benchmarks, like Mujoco, give the agent direct access to MDP state. Larger partially observable environments exist, such as Minecraft \citep{guss2021minerl}, but many of these require complex agent architectures and significant compute, even when considering lightweight versions like Craftax \citep{matthews2024craftax}. Consequently, isolating research questions and incremental algorithmic development becomes challenging. As a result, many CRL papers introduce a new toy CRL task, usually by taking a known environment and introducing some non-stationarity, and then conclude with a separate set of experiments in fully observable benchmarks \citep{lee2024slow,lee2023plastic,lyle2023understanding,lyle2024normalization,lyle2024disentangling,sokar2023dormant,nikishin2023deep,lyle2022understanding,elsayed2024addressing}.      

Work in CRL has largely focused on maintaining network trainability in the face of task non-stationarity but does not emphasize the role of partial observability. RL agents can both catastrophically forget and lose all ability to learn (so called loss of plasticity) \citep{abbas2023loss,dohare2024loss,anand2023prediction,lyle2024disentangling} when faced with either a sequence of tasks or some unobservable source of non-stationarity. 
Although loss of plasticity is widely observed, the precise cause remains elusive and a wide variety of potential culprits have been put forward including: dead and dormant neurons \citep{abbas2023loss, sokar2023dormant}, the beneficial properties of network initializations eroding over time \citep{kumar2020implicit,lewandowski2025learning,galashov2024non, dohare2024loss,anand2023prediction}, changes in target magnitude \citep{lyle2024disentangling}, and even ineffective hyper-parameter tuning \citep{mesbahi2025position}. The solutions, various forms of regularization, resetting, normalization, and special network architectures, all focus on maintaining plasticity, ensuring the agent can continually relearn in response to non-stationarity induced by task switches. Although task switching has been noted as a form of partial observability \citep{khetarpal2022towards}, to the best of our knowledge, prior work has focused on methods that enable continual learning of the policy and value function, but state-construction remains underexplored in CRL.       

In this paper we introduce a new testbed and set of experiments exploring an agent's ability to both track (continually update the value function and policy) and construct state. We introduce a lightweight environment generation testbed for continual learning called Forager\footnote{We use Forager as a testbed for continual
learning under partial observability, not as an ecologically realistic model of
foraging, and make no claims about optimal-foraging
theory~\citep{charnov1976optimal} or other models of patch
foraging~\citep{wispinski2022adaptive,fonseca2025world,davidson2019foraging,kilpatrick2021uncertainty}.
}. 
The agent's objective is to collect and avoid objects in a gridworld based on a limited, overhead field of view. 
Forager can run up to a hundred thousand frames per second with a constant memory footprint (as shown in Figure \ref{fig:jbw_grow}) giving it a significant advantage over other CRL benchmarks like Jelly Bean World (see Section \ref{related} for a discussion of related testbeds). Forager is parameterized by an adjustable field of view and world size, allowing careful control of the degree of partial observability. 
Forager is available in two implementations.
The CPU implementation allows running long-running experiments very quickly.\footnote{\href{https://github.com/steventango/forager}{https://github.com/steventango/forager}} The JAX implementation, Foragax, allows for massive parallelism for hundreds of independent trials in parallel on GPU.\footnote{\href{https://github.com/steventango/continual-foragax}{https://github.com/steventango/continual-foragax}} Foragax is useful for obtaining hyper-parameter sweeps and running a sufficient number of independent trials to achieve statistical significance. 
Finally, we specify several test cases as a starting point, including a variant that produces an infinite series of continual learning tasks that require both tracking and continual state construction. 

Our experiments reveal, unsurprisingly, that popular deep RL agents fail to learn continually or settle on suboptimal, static foraging policies. More interestingly, recently proposed continual learning approaches like regularization~\citep{kumar2024maintaining, ash2020warm},
CReLU activations~\citep{abbas2023loss}, and using multiple networks~\citep{anand2023prediction} do not help resolve the partial observability in Forager. Some methods appear to address loss of plasticity, however, they show limited success in continual learning through tracking. Whereas state construction methods, like simple memory traces~\citep{rafiee2023eye,janjua2023gvfs} and recurrent architectures~\citep{elelimy2024real} enable tracking but are still suboptimal compared to search baselines. The Forager benchmark suite and our extensive set of results provide a clear foundation for investigating continual learning in partially observable environments where both continual state construction and tracking are required, just like in the real world.

\section{Related CRL testbeds and benchmarks}
\label{related}
Currently available RL benchmarks are not well-suited to developing new CRL algorithms. Many of our large-scale benchmarks are not big worlds. The ALE environment is actually an MDP under-the-hood, which a recurrent agent should be able to perfectly simulate. The same is true of Mujoco, DM Control, and DMLab. There are several large-scale, realistic environments but these require long learning times and prohibitive computation, such as Minecraft~\citep{tessler2017deep,hafner2025mastering} and Causal World~\citep{ahmed2020causalworld}. NetHack~\citep{kuttler2020nethack} and Crafter~\citep{hafner2022benchmarking} are more computationally frugal variants, but remain quite complex and slow. Another approach is to take already expensive environments and switch between them, such as in Switching ALE~\citep{abbas2023loss}, Cora~\citep{powers2022cora}, and COOM based on Visual Doom tasks~\citep{tomilin2023coom}. These benchmarks are  computationally challenging as CRL experiments already necessitate longer training time to demonstrate successful learning or loss of plasticity.

Craftax ~\citep{matthews2024craftax}, substantially improves on Crafter but becomes impractical when execution is restricted to a single-stream of experience. A modified version of Craftax, ForageWorld, was recently used to benchmark foraging in an environment with food, water, and enemies ~\citep{simmons2025deep}. Interestingly, recurrent PPO was found to be highly effective, but required both significant environment parallelism and access to the privileged global (x, y) position of the agent via an auxiliary loss target. Although asynchronous training is widely used, support for single-stream RL is important for algorithmic progress \citep{abel2023definition, mesbahi2025position}. In addition, recent work has shown that the use of asynchronous training mitigates loss of plasticity~\citep{mayor2025the}, effectively masking one of the key challenges of CRL. 

Many papers introduce a bespoke, lightweight environment to highlight the challenges of continual learning. These environments are typically variants of supervised learning benchmarks or created by adding non-stationarity or task-switching to an existing RL environment. Examples include Up n Down \citep{nikishin2023deep}, image-classification MDPs ~\citep{lyle2023understanding}, and Slippery Ant \citep{dohare2024loss}. In many cases, these tasks are never used again in followup work.

\begin{wrapfigure}[15]{r}{0.27\textwidth}    
\vspace{-.2cm}
\centering
        \begin{subfigure}[b]{0.27\textwidth}
        \centering
        \includegraphics[width=\linewidth]{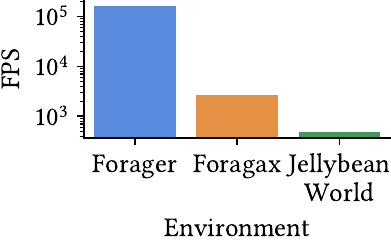}\\
    \end{subfigure}
    \begin{subfigure}[b]{0.27\textwidth}
        \centering
\includegraphics[width=\linewidth]{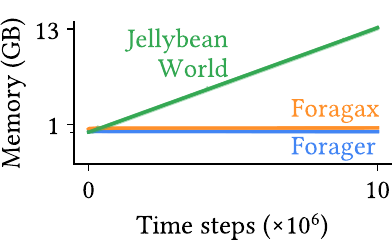}\\
    \end{subfigure}
  \caption{FPS and memory use of JBW and Forager.}
  \label{fig:jbw_grow}
\end{wrapfigure}
There are some environments designed to be big worlds. In Jelly Bean World (JBW), the agent navigates an infinite, procedurally generated gridworld collecting and avoiding objects. The agent's observation, like Forager, is an agent-centric, limited top-down field of view which, in principle, requires recurrent architectures or some other form of memory. Indeed, JBW was one of the few testbeds highlighted in \cite{khetarpal2022towards}'s survey. Yet, to the best of our knowledge, only two papers have used JBW ~\citep{anand2023prediction,mesbahi2025position}. One potential explanation is that JBW runs relatively slowly and has a memory footprint that grows as the agent explores, as demonstrated in Figure \ref{fig:jbw_grow}. In contrast, the Forager CPU implementation is significantly faster and the Forager JAX implementation, Foragax, enables running hundreds of independent trials in parallel on a single GPU. The better the agent becomes at continual learning, the more memory JBW uses and the slower it runs. As we discuss in the Supplement (Sections \ref{sec:benchmarking} \& \ref{jbw-like}), Forager does not use procedural generation, but still produces quantitatively similar results to JBW with a fixed memory footprint at a fraction of the costs, allowing for rapid prototyping of new ideas. 

AgarCL is a new benchmark based on the video game Agar.io that features multiple agents, changing dynamics, and a limited FOV~\citep{mohamed2026cell}. 
AgarCL remains an open challenging benchmark for continual reinforcement learning. In contrast to AgarCL's complex visual observations and hybrid action space, Forager deliberately uses simple grid world dynamics to achieve high simulation throughput to allow for rapid algorithmic development.

\section{Background}
In this paper, we consider continual reinforcement learning problems modeled as Markov Decision Processes (MDP)\footnote{We do not introduce the more involved Partially Observable MDP notation, because it is not needed for this paper.}
where certain information is not visible to the agent. The agent's interaction proceeds synchronously, where on each discrete time-step $t=1,2,...$ the agent chooses an action $A_t\in\mathcal{A}$ in state $S_t\in\mathcal{S}$.
The environment then, in part due to the agent's action choice, transitions to a new state $S_{t+1}$ and emits a scalar reward $R_{t+1}$. 
We assume that the environment is continuing, and performance is evaluated based on the average reward accumulated over the agent's lifetime.

We consider environments that use two approaches to limit agent observability. The first is to use non-stationary rewards, which is a form of partial observability because the dynamics producing the non-stationarity are hidden from the agent. 
A common approach to generate non-stationary rewards is to use a finite set of reward functions $\{R^{(0)},R^{(1)},...\}$ and every $\tau$ time-steps simply switch to a different reward function. Typically $\tau$, the switch events, and the set of possible rewards are not exposed to the agent. This setup can be framed as a multitask learning problem, but most researchers instead use such problems to measure how quickly agents can adjust to sudden, unexpected, unpredictable changes. Relatedly, the reward can be changed with time ($R_{t+1} \doteq r_t(S_t,A_t, S_{t+1})$),
 typically slowly, producing an infinite sequence of problems. Another approach to limit agent observability is to limit the agent's field of view. There are other ways to make the MDP partially observable necessitating continual learning, such as changing the transition dynamics according to some unobservable function, which we explore in our last experiment.

We run demonstrative experiments with DQN \citep{mnih2015human} and PPO~\citep{schulman2017proximal}. We largely make use of function approximation architectures based on feedforward, fully connected neural networks but also investigate recurrent agents with DRQN  ~\citep{hausknecht2015deep} and RTU-PPO~\citep{elelimy2024real}. 

\section{The Forager environment}
\label{forager}
Forager is a grid-world environment where the agent collects a variety of mushrooms with different rewards and respawning behaviors. The environment simulates an infinite world from the agent's point of view: instead of being a simple grid with boundaries, it is a torus where the agent can go in the cardinal directions indefinitely by wrapping around the edges. The agent has a limited field of view (FOV), making the environment partially observable. 
Forager is actually a family of environments, configurable via user specified parameters. %
With a smaller FOV, the environment is more partially observable and challenging. At the other extreme, the FOV can be set to the size of the world, providing full observability. Further, the placement and respawning of objects, reward functions, and colors in the environment can be configured and changed over time. 

\begin{wrapfigure}[11]{r}{.25\textwidth}
    \vspace{-0.5cm}
        \centering
        \includegraphics[width=\linewidth]{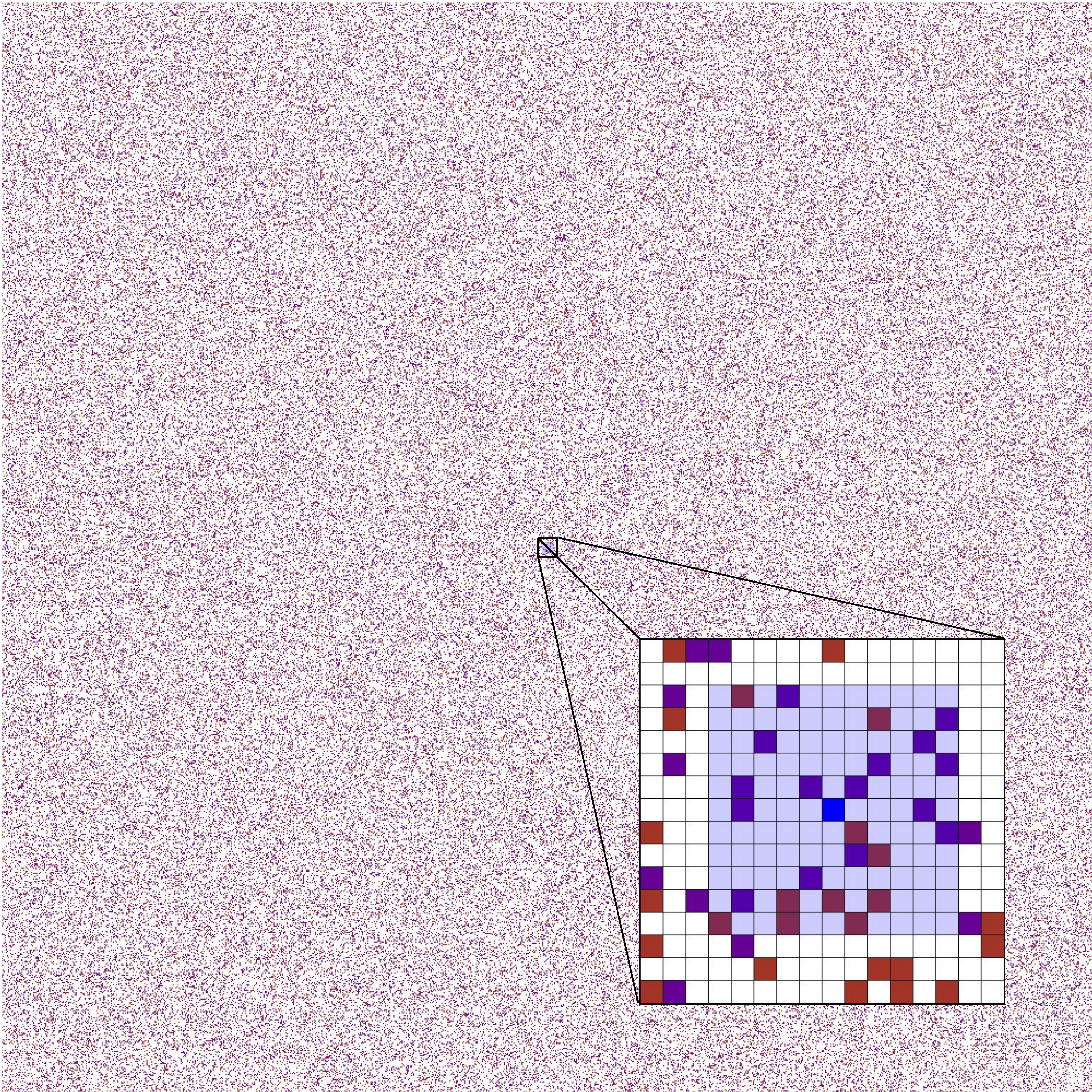}
        \caption{A sample foraging world.}
        \label{fig:forager_extra_large_env}
\end{wrapfigure}
Figure \ref{fig:forager_extra_large_env} shows an example Forager environment with two mushroom colors that can be collected by the agent. Figure \ref{fig:exp1} shows a different environment, where the agent forages for a variety of mushrooms, some high reward (dark brown), some lower reward (purple), and some poisonous and thus negative reward (yellow). The actions are \{\textit{up, down, left, right}\} and move the agent in the corresponding cardinal direction.  
The blue square represents the agent, and the light blue overlay represents its FOV. When the agent collects an item,
it disappears and the agent is rewarded. The objects either respawn in their original location or a random one according to user design. For example, in one of our experiments the high reward mushrooms respawn much more slowly than the other mushrooms, making them rare but more valuable. All other rewards are zero. Other collectible objects and unmovable ones (e.g., walls) can be easily added.

The observations are agent-centered, meaning that the agent is always at the center of the observation. The agent's observation is configurable to be either a RGB image of the agent's FOV or a binary tensor indicating the occupancy of each item type with shape (\textit{FOV}, \textit{FOV}, \textit{unique objects}). 

Non-stationary rewards are produced in multiple ways. One example is by decaying the reward for a mushroom on each step, simulating spoiling or rotting over time. Another is to have the value of a mushroom type change over time, simulating the agent's taste or nutritional needs changing over time. It might like one mushroom type in early life, and later in life, start preferring another. These non-stationary rewards would encourage agents to continually adapt to their changing environment.

\section{Desiderata for a continual learning testbed}
In this section we review the design principles of JBW~\citep{platanios2020jelly}, the characteristics of good continual learning benchmarks~\citep{khetarpal2022towards}, and the properties of continual learning problems motivated by a recent formal definition of continual learning~\citep{abel2023definition}.

The JBW environment predates most recent interest in CRL and was itself largely inspired by the work on the Never-ending Language Learning system~\citep{carlson2010toward}. The aforementioned system crawled the internet for years, reading and understanding webpages. The designers of JBW were inspired by the idea of unbounded learning systems, but also wanted to build a highly configurable environment where new experiments could be setup and run quickly. JBW is designed to require never-ending learning using two sources of partial observability: (1) a limited FOV in an infinite, procedurally generated world to navigate, and (2) non-stationary reward functions that either change slowly or periodically change completely. 
Forager was designed with the same principle, and differs in that it is more scalable than JBW and does not include some of the advanced observation features like multi-modal observations (i.e., scent) and object occlusion.

Due to the lack of appropriate continual reinforcement learning benchmarks, others have proposed a set of properties that future benchmarks should have \citep{khetarpal2022towards}. We mention them here as Forager does not achieve all these properties in full. In particular, \cite{khetarpal2022towards} identify the following: 1) learning should be evaluated online and incrementally, 2) good performance should require discovery and composition
of skills, 3) some form of physics which the agent can model, and 4) learnable causal dynamics such as object interactions. While our experiments with Forager do not demonstrate skill discovery and composition, it is not hard to imagine that skills for navigating to the edge of the FOV or the nearest mushroom would be useful for planning and acting. We only explore a few object types in this paper (consumable objects and walls), but more complex affordances like pushing or carrying objects could be easily added. Forager certainly does not implement interesting physics. We believe this is not worth the potential computational penalty and our experiments show Forager is a significant challenge for current algorithms. 

Another good way to design an environment useful for continual learning is to look at how it matches formal notions of continual reinforcement learning. \cite{abel2023definition} provides a formal problem specification that intuitively translates to problems where the best agents do not converge. In the stationary variants of Forager, even with a limited FOV, there are agents that can converge. However, it is a hard memory problem and for the limited agents we consider, it is likely necessary to track (i.e., learn forever). In the non-stationary variants of Forager, such as the one 
requiring never-ending learning 
(described below), the agent cannot model the partial observability and must track. Forager, therefore, is a continual learning problem according to this definition.

Forager largely achieves the desiderata laid out in the literature, just as JBW did before it. The key differences are motivated by the need for (1) a more lightweight environment and (2) to remove design features that are not critical to challenge current algorithms.

\section{Investigating the impact of field of view}

Forager provides an easy way to modify the degree of partial observability by altering the FOV observed by the agent. 
With a limited FOV, the agent must be able to use memory (or recurrent state updates) to remember the locations of rewarding mushrooms in the world beyond its FOV, and also predict when mushrooms will respawn outside of its FOV. In this section, we provide a simple demonstration of the effect of FOV. %

  \begin{wrapfigure}[28]{r}{0.4\textwidth}   
\vspace{-0.7cm}
\centering
\begin{subfigure}[b]{0.2\textwidth}
        \centering
       \setlength{\fboxsep}{0pt}
       \fbox{\includegraphics[width=\linewidth]{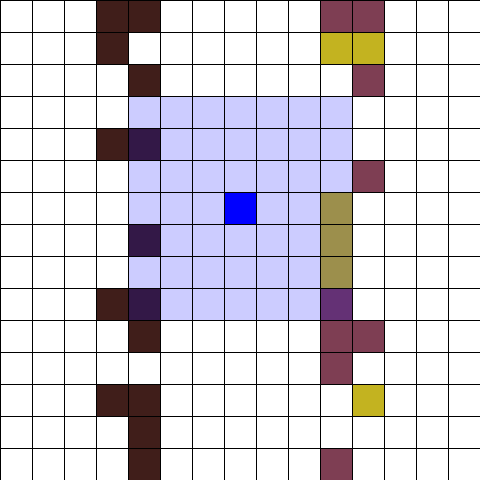}}
    \end{subfigure}
        \begin{subfigure}[b]{0.35\textwidth}
        \centering
  \includegraphics[width=\textwidth]{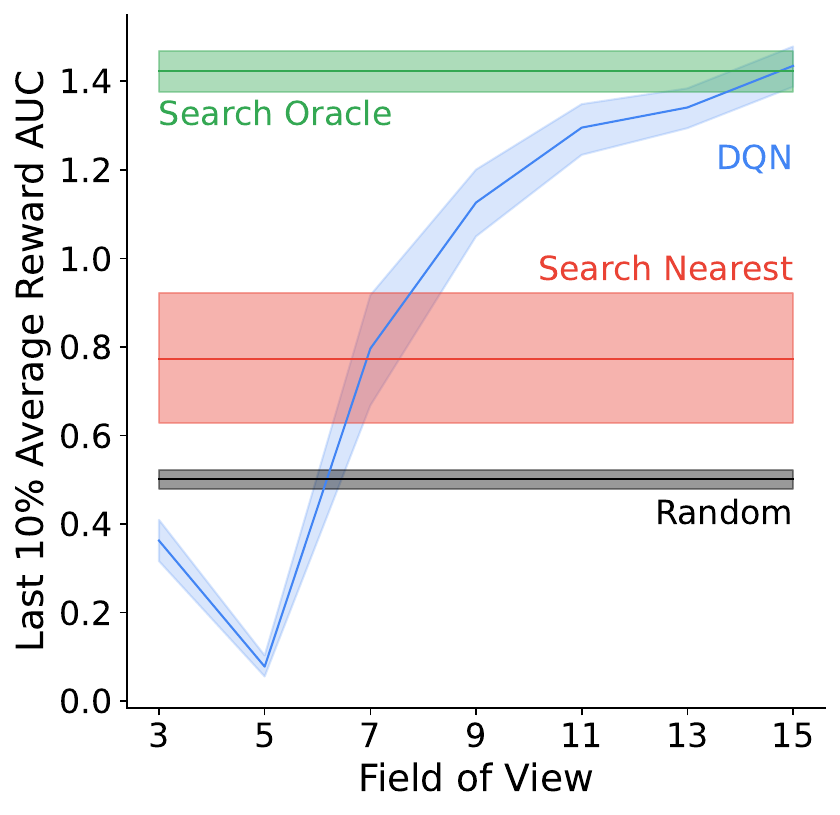}

    \end{subfigure}
\caption{
A simple Forager environment (ForagerTwoBiome-v1) with two biomes. The light blue overlay represents the FOV. The other colored squares are mushrooms which generate reward when collected. On the bottom we see how decreasing FOV makes the task more challenging for DQN. Results are averaged over 30 independent trials; shaded regions are $95\%$ bootstrap CI.}\label{fig:exp1}
\end{wrapfigure}
The environment is visualized in Figure \ref{fig:exp1} (top). The idea is inspired by biomes, regions in the natural world where different flora and fauna live. After a foraging agent consumes all the valuable flora in one biome, it must navigate to the next one and continue foraging. This is made more challenging because the flora in different biomes do not regrow at the same rate. In this environment the agent is collecting mushrooms. The mushrooms on the left, morels, are high value (+30 reward), but respawn very slowly compared to the mushrooms on the right, which give rewards of +1 for oyster (purple) and -1 for deathcap (yellow) mushrooms. The mushrooms on the right respawn much more quickly. A smart agent should collect the morels whenever they are available and otherwise focus on collecting oyster mushrooms and avoiding deathcaps; continually switching back and forth.

In Figure \ref{fig:exp1} (bottom) we see that as the FOV size decreases, the environment becomes more difficult for DQN. With the largest FOV sizes, DQN performed similarly to the Search Oracle. From videos of agent behavior,\footnote{\url{https://sites.google.com/view/rlc2026-forager/}} we see the expected behavior: DQN with larger FOV sizes acts almost exactly like the Search Oracle: foraging the oyster mushroom biome mostly, but periodically crossing the gap to forage the morels biome when they respawn. As the FOV size decreased, DQN performance deteriorated. For FOV size 7, DQN resembled Search Nearest in behavior, basically exclusively foraging oyster mushrooms and only rarely visiting the morel biome. The FOV sizes of 3 and 5 do not allow the agents to see across the gap. These agents performed worse than the baselines and stumbled around the world lacking reward-seeking behavior.

The results from this experiment were exactly as expected. With a large enough FOV, the environment is fully observable to the agent and DQN should and does learn near-optimal foraging. As we reduce the FOV, the agent cannot see when the morels have respawned, and therefore DQN would need to learn to take excursions into the left biome to check if the morels had respawned. We saw no evidence of this behavior emerging with smaller FOV sizes, which, again, is not surprising because this would require remembering the location of previously rewarding locations in the grid.

\section{Investigating never-ending relearning in Forager}
We expect a big world to require continual learning, exploration, and state construction from agents. The previous experiment does not actually require continual, unending learning, because everything is observable to the agent given enough interaction with the world---there is no hidden state-transition mechanism. 
In this section, we explore a switching-task variant of Forager that requires unending policy change (for non-recurrent agents) to study the impact of recently proposed continual learning mitigation strategies. Then we investigate how state-construction methods support continual learning under constant task switching.   

\begin{figure}[tb]
    \centering
        \begin{subfigure}[t]{0.11\textwidth}
        \centering
        \caption{}
        \label{fig:PPO-DQN-a}
       \setlength{\fboxsep}{0pt}
       \vspace{.2cm}
        \fbox{
        \includegraphics[angle=90, width=0.9\linewidth]{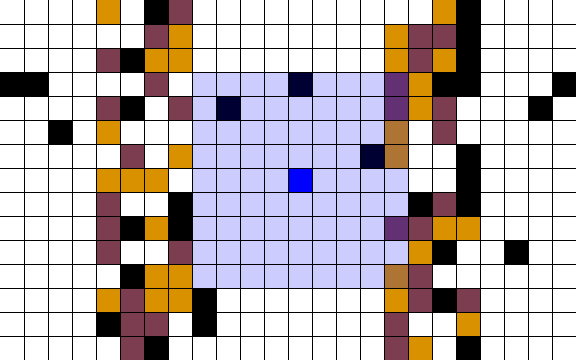}}
    \end{subfigure}
    \begin{subfigure}[t]{0.26\textwidth}
        \caption{}
        \label{fig:PPO-DQN-b}
        \includegraphics[width=\linewidth]{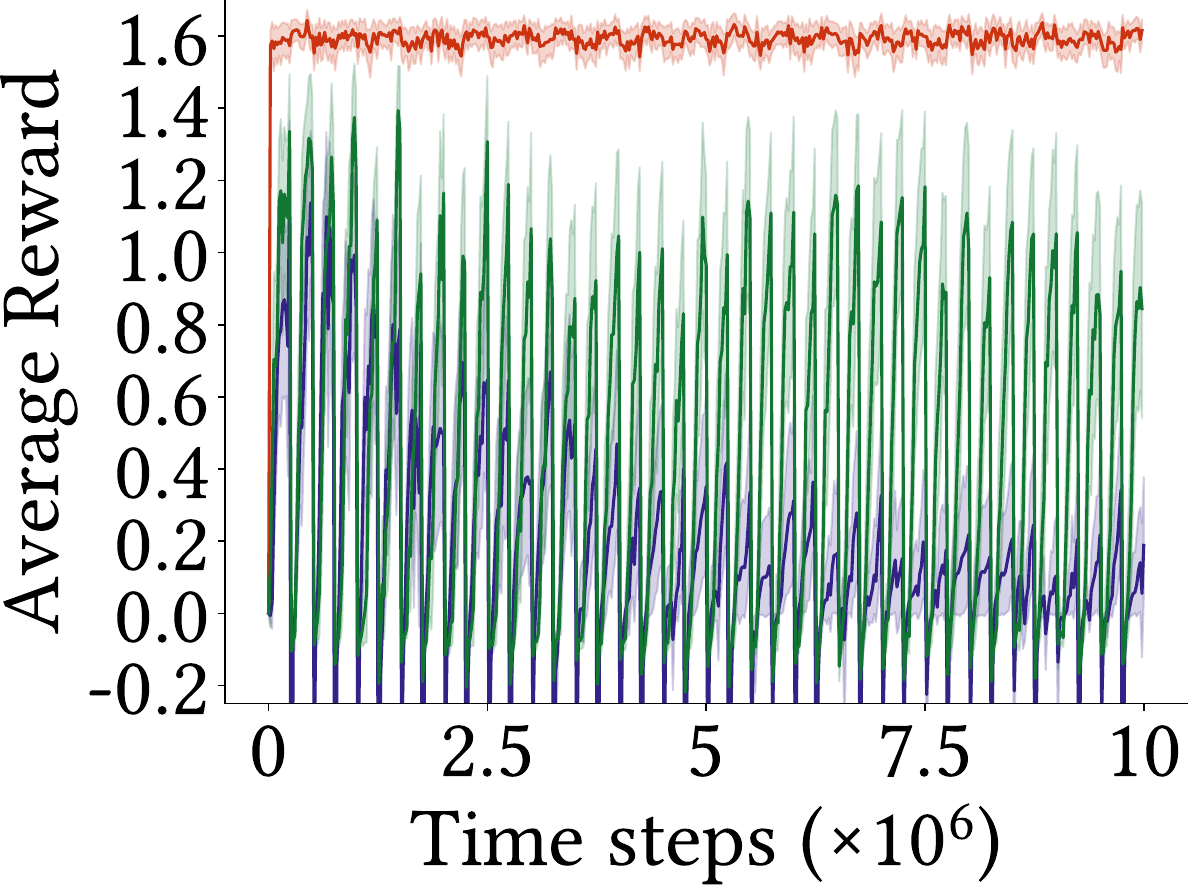}
    \end{subfigure}
    \begin{subfigure}[t]{0.26\textwidth}
            \caption{}
        \label{fig:PPO-DQN-c}
        \includegraphics[width=\linewidth]{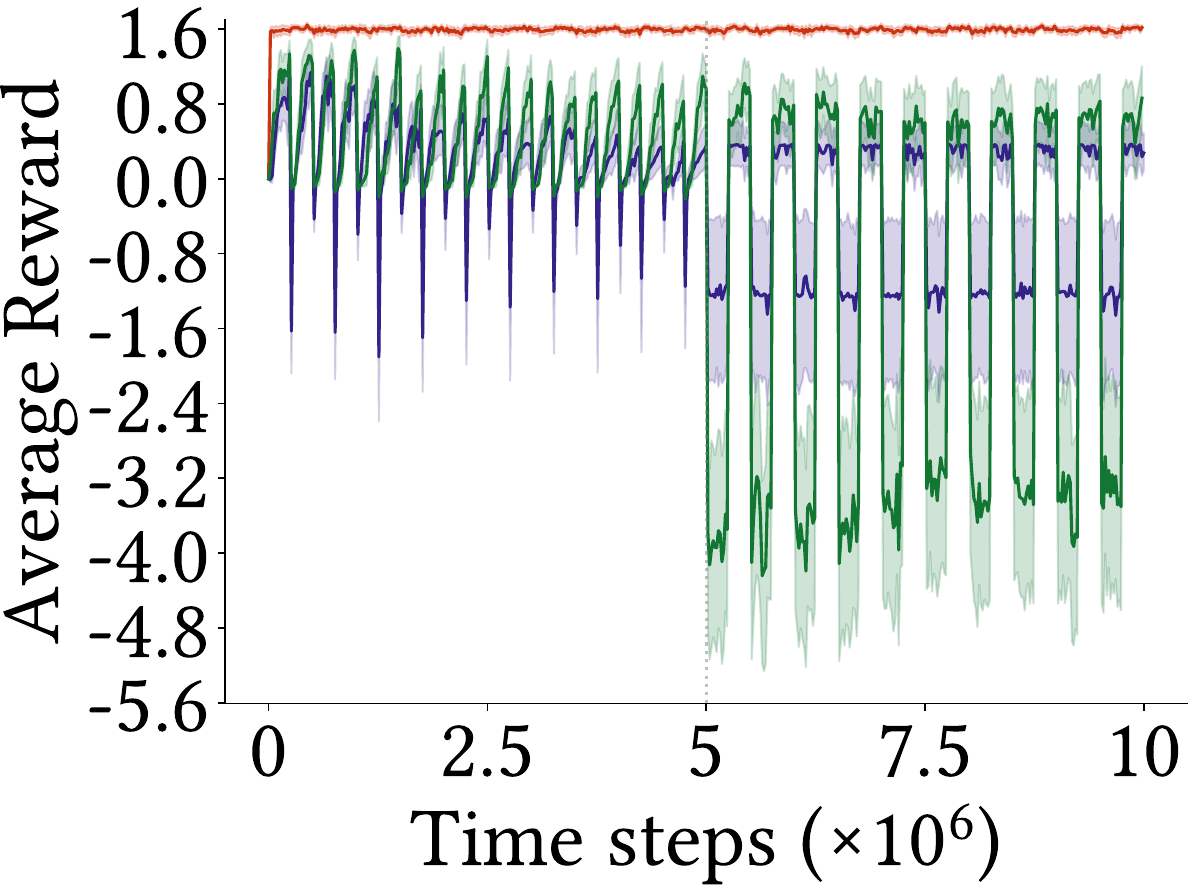}
    \end{subfigure}
    \begin{subfigure}[t]{0.3\textwidth}
            \caption{}
        \label{fig:PPO-DQN-d}
        \includegraphics[width=\linewidth]{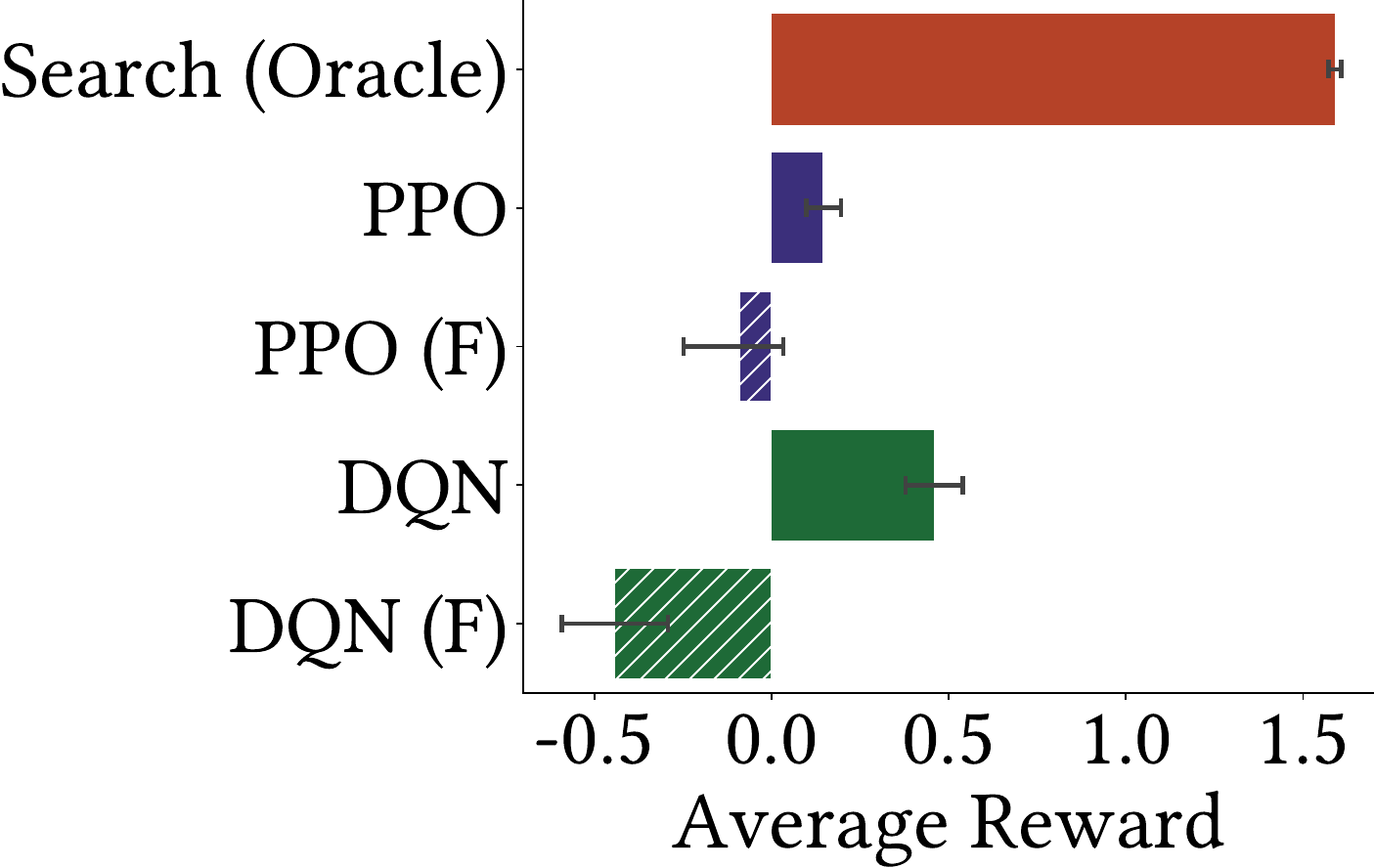}
    \end{subfigure}
    \centering
    \caption{The performance, both learning curves and summary bar charts, of DQN and PPO. The bar chart serves as a legend for the learning curves. \textbf{(a)} Visualization of continual relearning task (ForagaxNeverEndingRelearning-v1). \textbf{(b)} We see both agents fall significantly short of the search baseline. DQN exhibits minor degradation over time, whereas PPO exhibits dramatic loss of plasticity. \textbf{(c)} However, by freezing learning halfway through the experiment, we see further drop in performance indicating both agents were indeed learning (poorly) to adapt to task switches. \textbf{(d)} The hatched bars report the performance of the agents frozen after 5 million steps, labeled (F).} 
    \label{fig:PPO-DQN}
\end{figure}

The setup of this experiment is similar to the previous 2-biome experiment, where there are two regions of interest with collectible mushrooms, but which mushrooms the agent should collect changes periodically and is region specific. The reward associated with each color is different on each side and this mapping changes every 250{,}000 steps according to a hidden schedule. In the beginning, the agent should collect purple mushrooms (+4 reward) and avoid orange (-2 reward) in the top biome and all mushrooms are negatively rewarding in the other biome (-8 for purple and -14 for orange). After the switch, all the mushrooms generate negative rewards in the top biome (-14 for purple and -8 for orange), but one of the mushroom types in the bottom biome generates positive reward and the agent should collect orange (+4) and avoid purple (-2) there. Figure \ref{fig:PPO-DQN-a} provides a screen shot of the environment. This environment includes impassible and unmovable walls both to obstruct the agents and to help agents localize themselves. Since the mushrooms in both biomes are visually indistinguishable to the agent with a limited FOV (9), most agents are forced to constantly relearn their foraging policies after each switch. A recurrent agent could, in principle, learn to anticipate the switch and navigate between biomes more effectively. To help focus on investigating the continual learning aspect of this formulation, we one-hot encoded each color as the input to agents. In addition, all agents receive last reward and last action as part of their inputs.

The first question is can our non-recurrent deep RL methods continually learn in this setup? We tested DQN and PPO as representative value-based and policy gradient algorithms. For the remaining experiments, we tune the hyperparameters of all learning agents using 10\% tuning with 10 independent trials, as tuning for the entire lifetime of the agent is not well aligned with the goals of CRL \citep{mesbahi2025position}. In Supplement  \ref{sec:experiment_details}, we outline all the hyperparameter ranges swept for each experiment. To contextualize the performance, we include a Search Oracle baseline that has full access to the underlying state (the precise location and reward for all objects in the world).%

Figure \ref{fig:PPO-DQN-b} summarizes the results. All results are averaged over 30 independent runs and we report $95\%$ bootstrap confidence intervals. We see DQN generally outperforms PPO, but does not reach the performance of the oracle, but as you will see later some learning methods perform much better than this. DQN appears to be suffering from minor loss of plasticity, while PPO is dramatically getting progressively worse with more task switches. Both DQN and PPO were indeed relying on continual learning. To see this clearly, we froze learning after 5 million steps in Figure \ref{fig:PPO-DQN-c}. We see a drop in performance of both agents; the frozen agents could only effectively collect mushrooms on whatever side they were frozen on.

\section{Do mitigations help in Forager?}

Next, we compared the impact of different mitigation strategies that have recently been proposed to promote CRL and prevent loss of plasticity. Since there are a large number of such mitigations, we investigate representatives from several categories, including: (1) regularizing the weights towards the network initialization \citep{kumar2024maintaining}, (2) the CReLU activation that prevents neuron death ~\citep{abbas2023loss}, (3) Shrink \& Perturb ~\citep{ash2020warm}, (4) L2 regularization ~\citep{dohare2024loss}, (5) Permanent-Transient Q-learning (PT-DQN) \citep{anand2023prediction} which uses multiple Q-functions to distill previous learning, and (6) ReDo, which resets dormant neurons ~\citep{sokar2023dormant}. All methods, including base DQN and PPO agents, include layer normalization, which appears to be beneficial for preventing plasticity loss ~\citep{lyle2024normalization}. We restrict our focus to mitigations proposed for RL settings and exclude others that require task labels and task boundaries, like EWC ~\citep{kirkpatrick2017overcoming}, as we are interested in approaches that can learn in the face of unexpected task changes. 

\begin{figure}[htb]
    \centering
        \begin{subfigure}[t]{0.49\textwidth}
            \caption{}
        \label{fig:mitigations-a}
        \includegraphics[width=\linewidth]{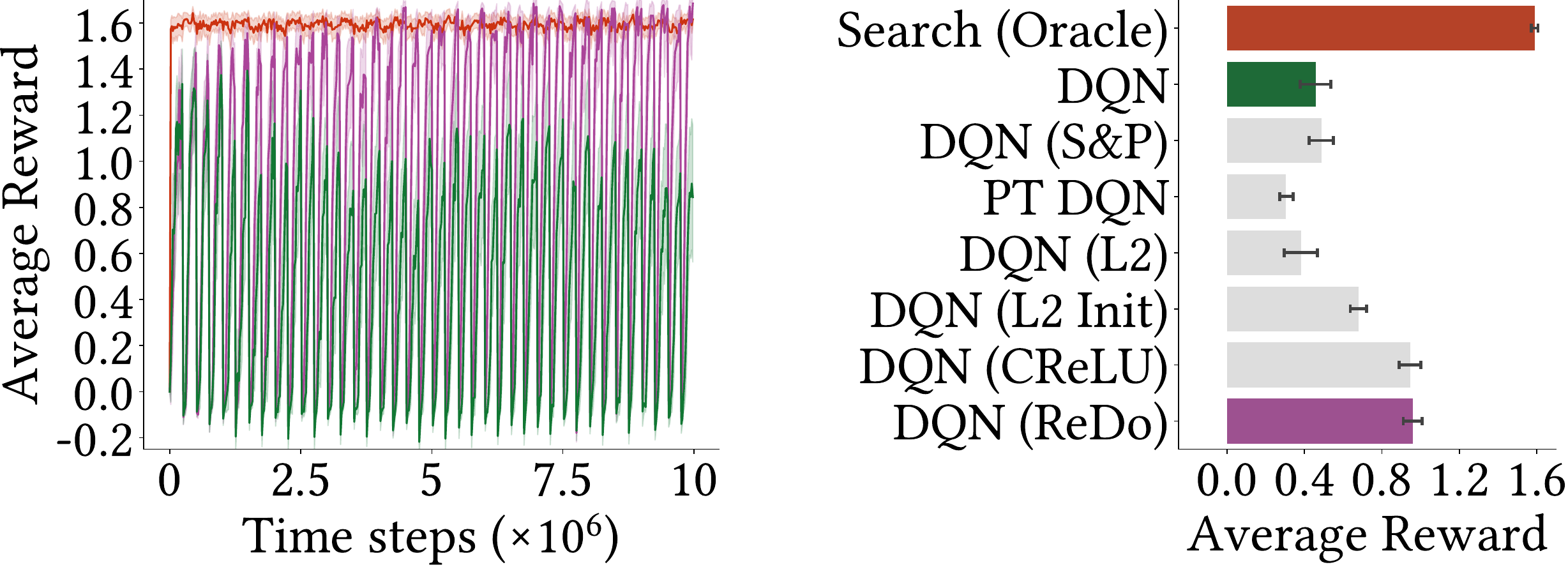}

    \end{subfigure}
        \begin{subfigure}[t]{0.49\textwidth}
            \caption{}
        \label{fig:mitigations-b}    
        \includegraphics[width=\linewidth]{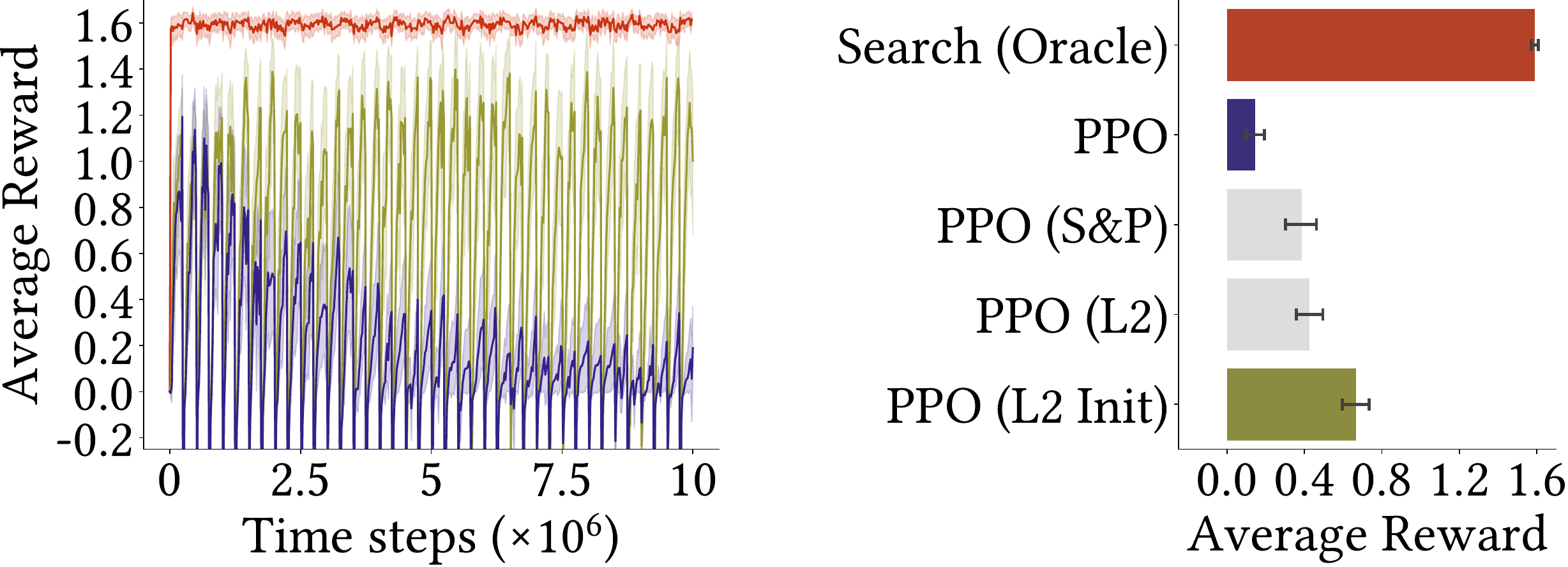}

    \end{subfigure}
    \centering
    \caption{The impact of loss of plasticity mitigation strategies on DQN and PPO. \textbf{(a)} For DQN, only regularizing the weights to stay close to initialization (L2 Init), CReLU activations, and ReDo significantly outperform the base algorithm. \textbf{(b)} For PPO, Shrink \& Perturb, L2, and L2 Init all help and the latter appears to prevent loss of plasticity. The learning curves only include the best mitigations to reduce visual clutter. }
    \label{fig:mitigations}
\end{figure}

Figure \ref{fig:mitigations-a} shows the impact of these mitigations when combined with DQN and Figure \ref{fig:mitigations-b} reports similar results with PPO. Note that 1) PT-DQN and ReDo were first demonstrated with DQN, so we did not extend them to PPO and 2) we did not test PPO with CReLU as PPO is almost exclusively used with Tanh. DQN's performance was significantly improved by CReLU, L2 Init, and ReDo, but not by PT-DQN, L2, nor Shrink \& Perturb. It is worth noting that DQN did not exhibit a severe loss of plasticity and the mitigations mostly increased how quickly DQN adapted to task switches. For PPO, L2 Init completely mitigated loss of plasticity, L2 and Shrink \& Perturb helped largely mitigate it. Across both algorithms, only L2 Init provided consistent benefit, though none of the mitigations provided sufficient benefit to get close to the Search Oracle.

\begin{figure}[tb]
    \centering
    \begin{subfigure}[t]{0.27\textwidth}
            \caption{}
        \label{fig:simple_memory-a}        \includegraphics[width=\linewidth]{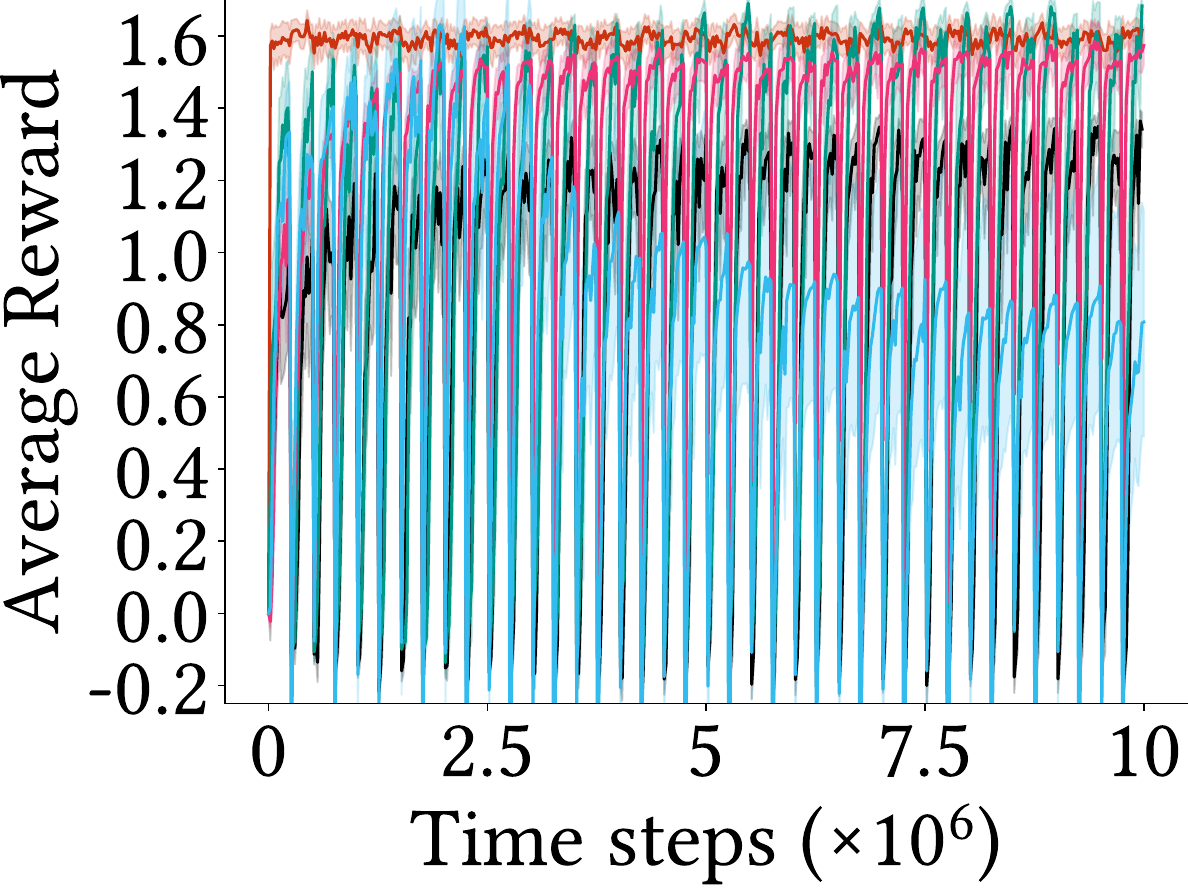}

    \end{subfigure}
    \begin{subfigure}[t]{0.27\textwidth}
            \caption{}
        \label{fig:simple_memory-b}        \includegraphics[width=\linewidth]{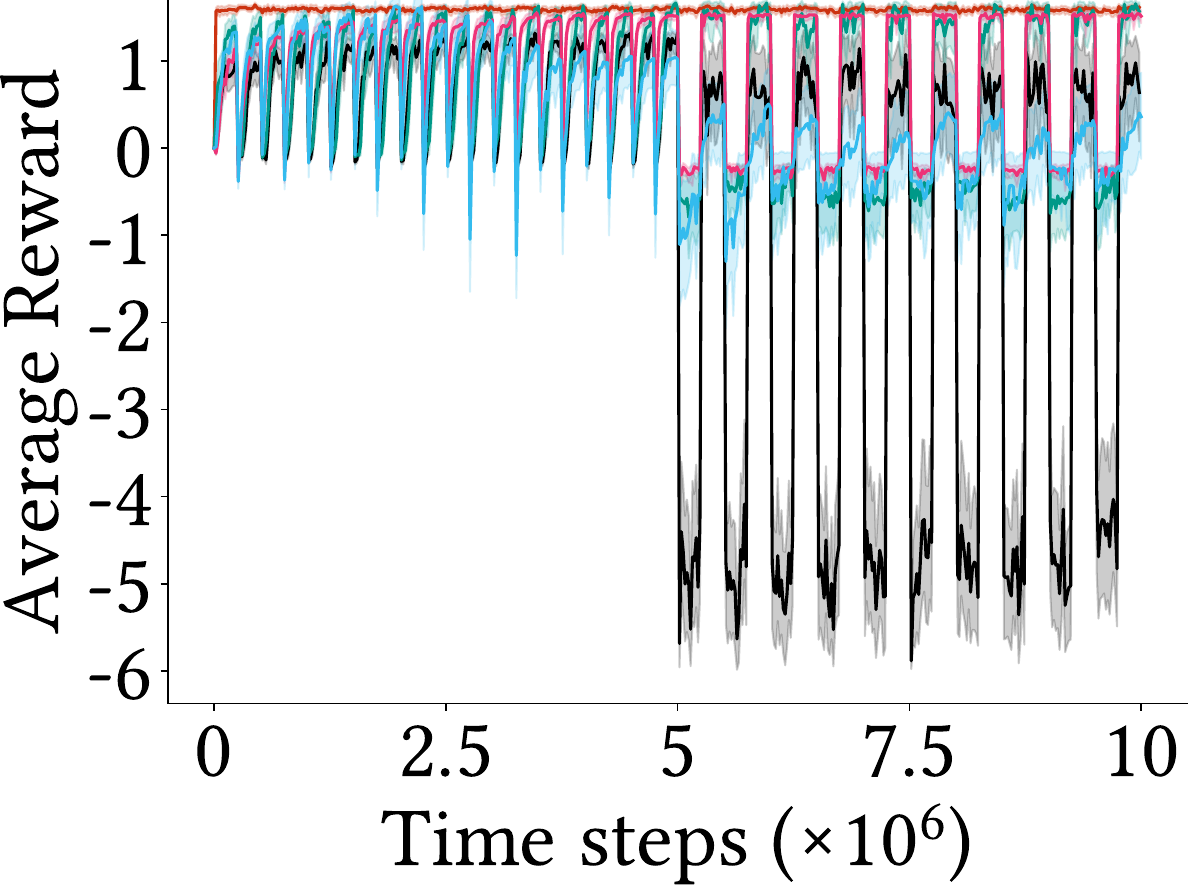}

    \end{subfigure}
    \begin{subfigure}[t]{0.4\textwidth}
            \caption{}
        \label{fig:simple_memory-c}        \includegraphics[width=\linewidth]{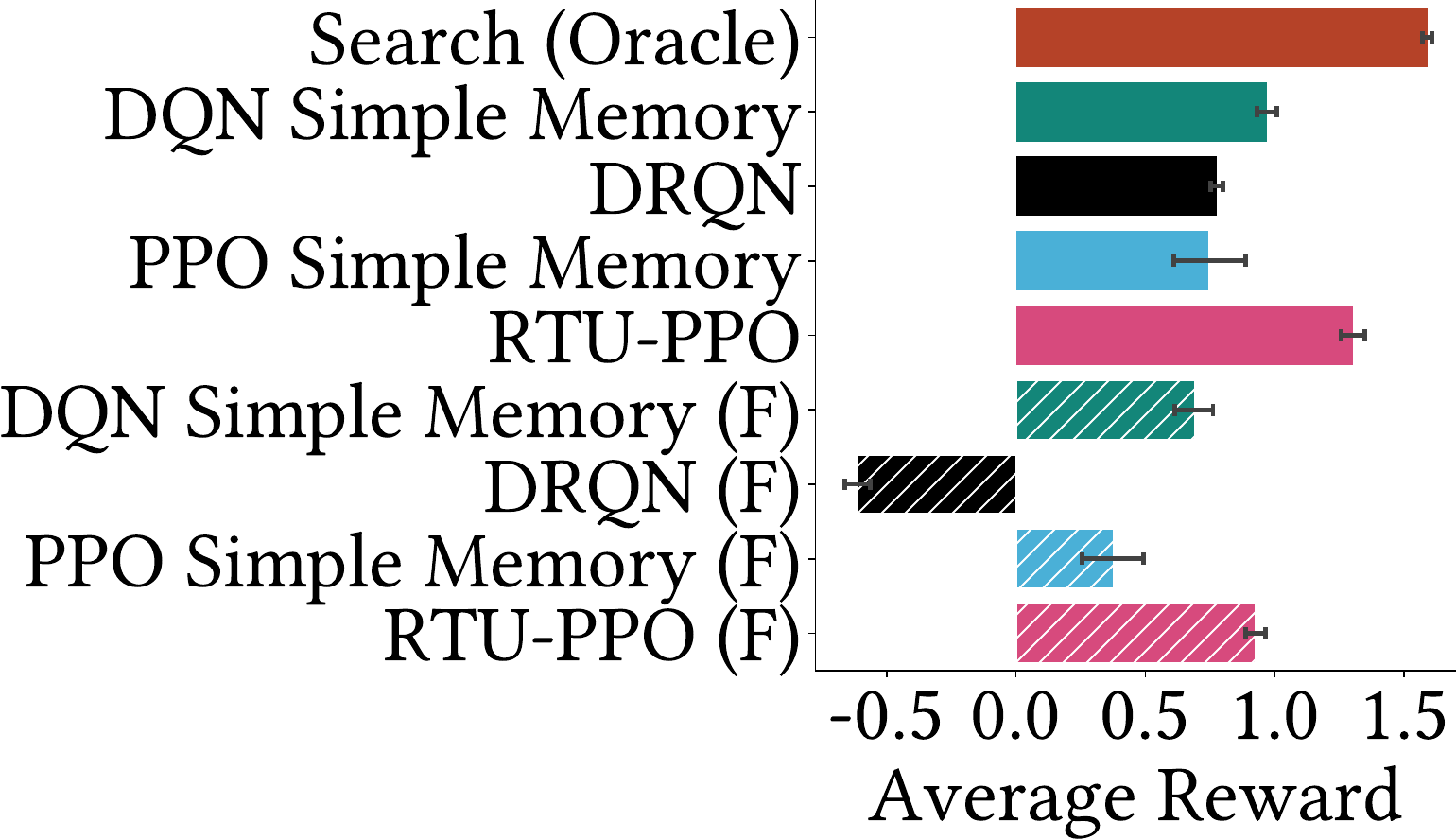}

    \end{subfigure}
    \centering
    \caption{The performance of state construction methods in Forager.
    \textbf{(a)} We see RTU-PPO outperforms all other learning methods in this environment, DRQN outperforms DQN, and both DQN and PPO with a simple reward trace do well. 
    \textbf{(b)} We see the impact of freezing these agents halfway through the experiment. All agents perform worse after freezing---indicating that they were indeed continually learning at the point of freezing. 
    \textbf{(c)} Interestingly, the frozen performance of RTU-PPO and both simple memory approaches is better than all other frozen methods indicating these methods have extracted some additional state information that makes the policy less dependent on tracking. 
    }
    \label{fig:simple_memory}
\end{figure}

\section{On the need for state-construction in CRL}
The previous results are in some sense a mixed bag. PPO demonstrates loss of plasticity---the longer it learns, the worse it gets. DQN does not suffer in the same way. PPO performed significantly worse than the search baseline and, as you will see next, it is possible to perform better. Finally, while the mitigation strategies reduced loss of plasticity, they still underperformed the search baseline.

Forager, like any big world, requires both continual policy and value learning and state construction to do well. The switching structure certainly necessitates periodic relearning of the policy, but an agent with memory could do more. The limited FOV means there is significant aliasing in the agent's location, making navigating between biomes difficult. In addition, memory would help the agent anticipate the switches rather than relying on sampling  negative rewarding mushrooms before adapting. 
While the mitigations in the previous section reduced loss of plasticity and slightly improved performance, we show that state construction  not only mitigates loss of plasticity but also more effectively supports continual learning.

To highlight this we augmented the base agents with simple memory structures. We included
an exponentially weighted memory trace ~\citep{rafiee2023eye} of the recent rewards as input observations to DQN and PPO. Comparing Figures \ref{fig:simple_memory-a} and \ref{fig:simple_memory-c} with Figure \ref{fig:PPO-DQN-d}, both agents exhibit significant performance improvements with the addition of simple memories of prior rewards.
PPO with a simple memory still exhibits loss of plasticity but much more slowly than before. DQN with a simple memory at times matches the Search Oracle and equals the best mitigation, ReDo. In addition, comparing Figure \ref{fig:PPO-DQN-d} with Figure \ref{fig:simple_memory-c}, the frozen policy from DQN with simple memory outperforms the frozen policy from vanilla DQN.

Exponential memories are extremely limited, so we also investigated the impact of recurrent variants of DQN and PPO. We tested a modified variant of the DRQN algorithm \citep{hausknecht2015deep} which combines DQN with an RNN trained via truncated backprop through time (TBPTT). We modernized DRQN to use gated recurrent units (GRU) and stored the hidden state in the buffer and employed a hidden state warm up to mitigate the impact of stale states in TBPTT similar to \cite{kapturowski2018recurrent}. We also include a recent recurrent PPO algorithm with an RTU layer that uses a diagonal approximation and a sin-cosine encoding to allow fast realtime recurrent learning updates~\citep{elelimy2024real}. 

The results are fairly striking. 
DRQN improves the performance of DQN but still falls short of the simple memory approach and RTU dramatically improves the performance of PPO. Loss of plasticity is eliminated and this agent achieves the best performance of all learning agents---better than DQN with ReDo---very close to the search baseline with low variance in performance. This exceptional performance is not because RTU-PPO has extracted the underlying state and converged to a static high performance policy. As we see in Figure \ref{fig:simple_memory-b}: RTU-PPO when frozen exhibits a significant performance drop, thus providing clear evidence that this agent is indeed continually relearning its policy of every switch, in addition to achieving better performance due to the recurrent network mitigating partial observability.

\section{An unending sequence of foraging tasks}
\begin{figure}[bt]
    \centering
    \begin{subfigure}[t]{0.18\textwidth}
        \centering
        \caption{}
        \label{fig:forager_big_v5-a}
        \setlength{\fboxsep}{0pt}
        \fbox{\includegraphics[width=\linewidth]{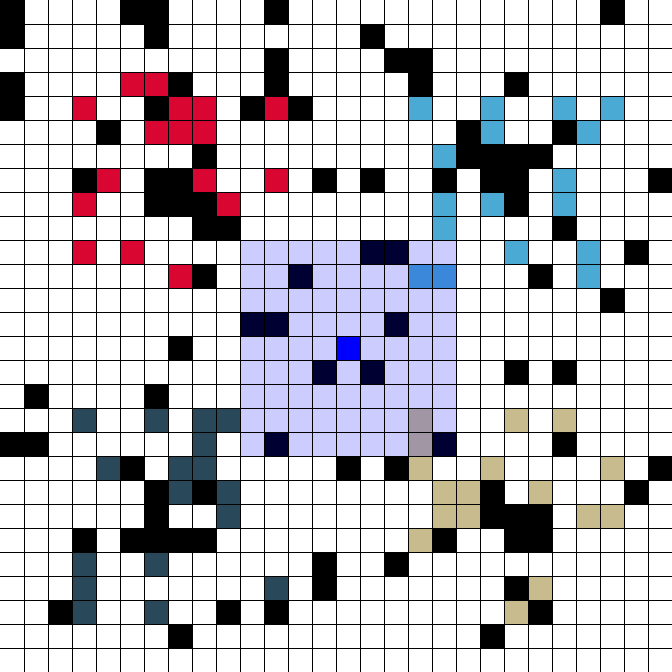}}
        \vspace{1.2em}
    \end{subfigure}
    \hfill
    \begin{subfigure}[t]{0.3\textwidth}
        \centering
        \caption{}
        \label{fig:forager_big_v5-c}
        \includegraphics[width=\linewidth]{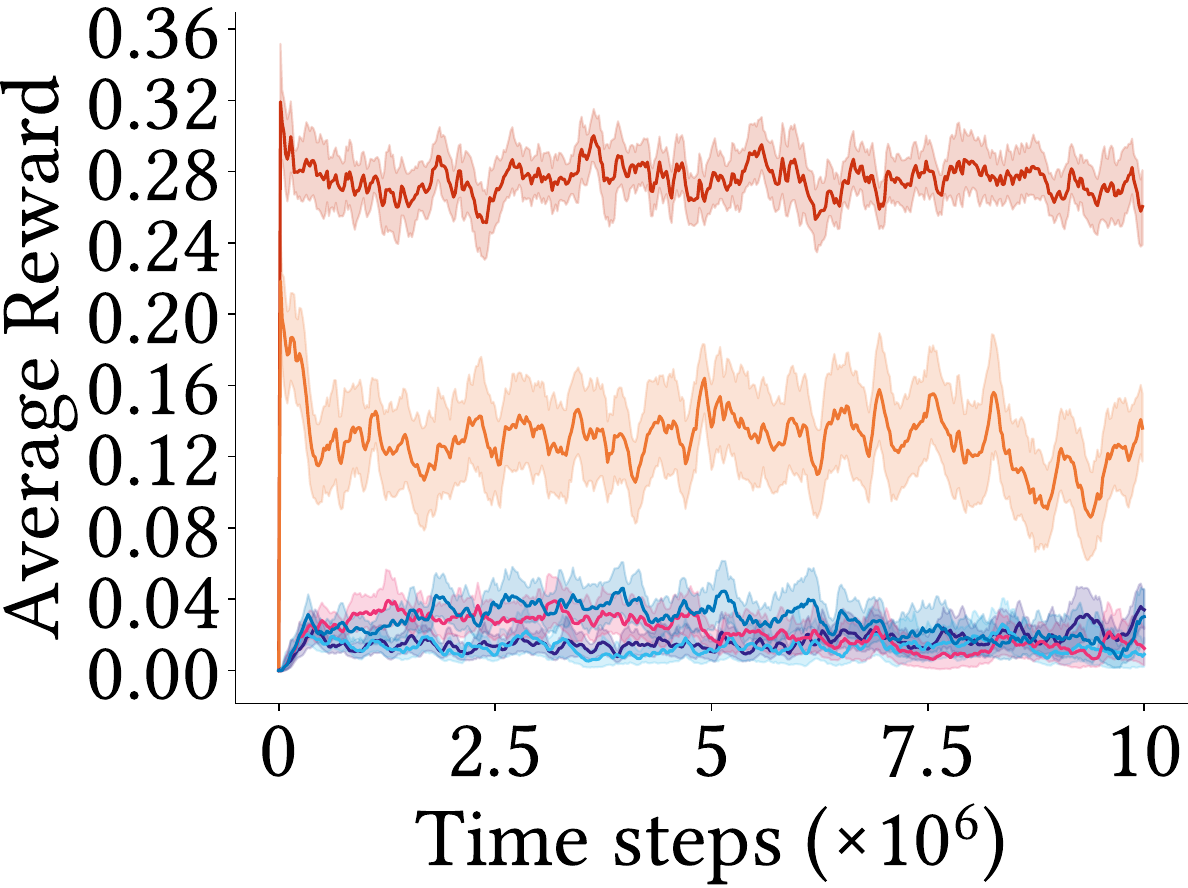}
    \end{subfigure}
    \hfill
    \begin{subfigure}[t]{0.4\textwidth}
        \centering
        \caption{}
        \label{fig:forager_big_v5-d}
        \includegraphics[width=\linewidth]{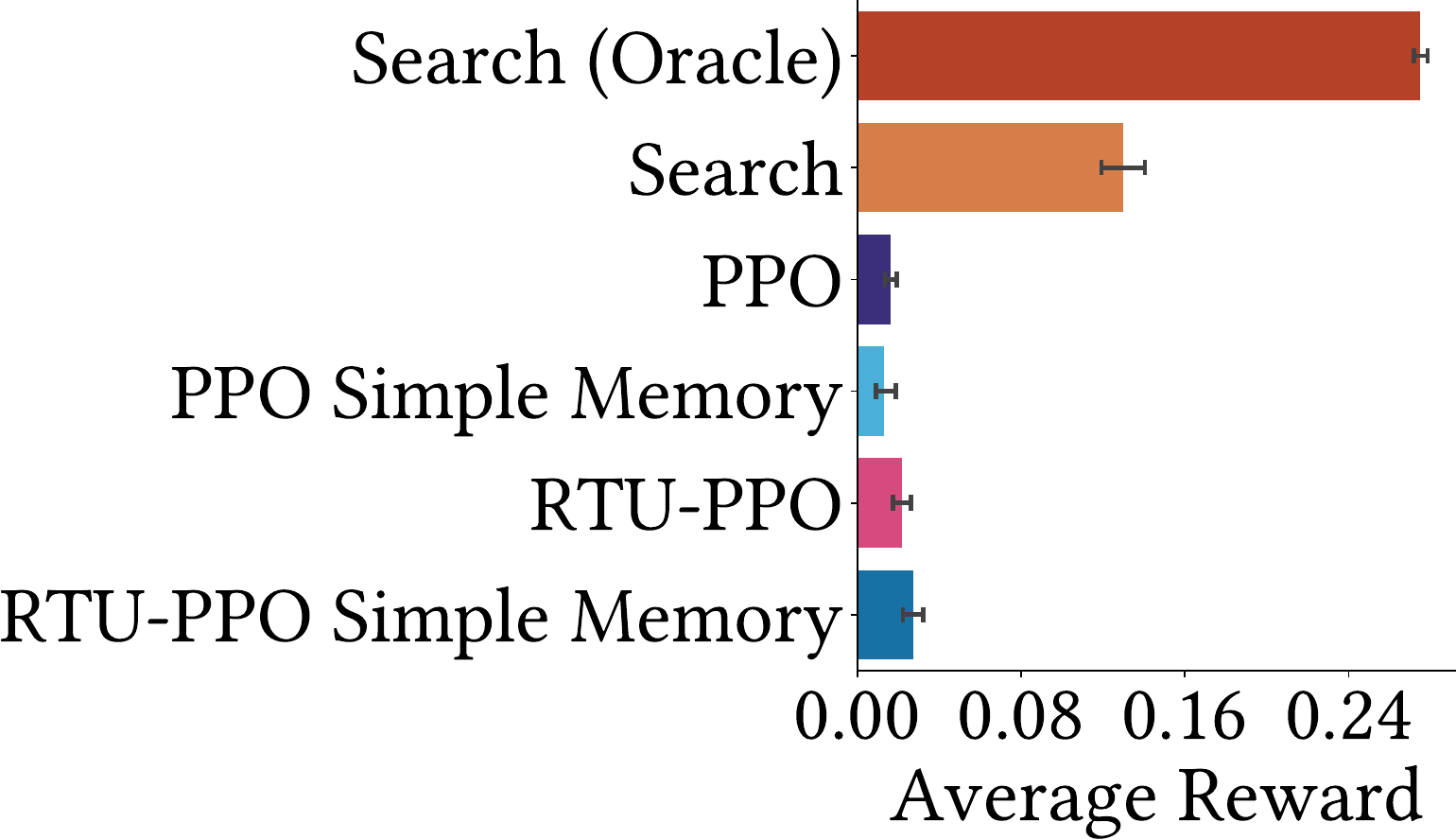}
    \end{subfigure}
    \caption{Foraging experiment with an unending series of tasks  (ForagaxUnendingTasks-v1).
    \textbf{(a)} Environment visualization: the blue square is the agent; the red, light blue, dark gray, and beige squares are different mushroom species; and the black squares are walls.
    \textbf{(b)} The learning curves of several learning agents.
    \textbf{(c)} The average reward obtained by each method. RTU-PPO Simple Memory combines both RTU and simple memory approach. The Search agent performs local search within the same FOV as other agents. The Search Oracle agent has access to the entire grid. Both search agents have privileged information about the current reward. 
    All results are averaged over 30 independent trials and the shaded regions and error bars represent $95\%$ bootstrap CI.}
    \label{fig:forager_big_v5}
\end{figure}

In this section, we introduce a challenge problem for CRL: an environment with an unending sequence of foraging tasks which change based on the agent's actions and unobservable timeseries. 
As visualized in Figure \ref{fig:forager_big_v5-a}, there are four biomes, each with its own mushroom species. Each mushroom species has a randomly sampled color and reward function that fluctuates with time based on a unique time series described in Supplement \ref{sec:unending}. Every 100 steps in the environment, a global cue signals the biome with the best rewarding mushrooms for 10 steps. The task switches are driven by agent behavior. 
When an agent consumes 10,000 individuals of a mushroom species, that species goes extinct and is replaced by a new species with a new color and new reward, forcing the agent to continually adapt its foraging strategy while also remembering the cue.

We focus our evaluation on PPO-based agents, as the previous experiments demonstrated that RTU-PPO outperformed DQN-based agents and other mitigations.
We use a convolutional neural network to handle RGB input. 
Additional experiment details are available in Supplement \ref{sec:unending}.

Figure \ref{fig:forager_big_v5} showcases the results of this experiment. The performance of all PPO variants plateaued early and did not improve further over the 10 M steps of training. All learning agents were unable to reach the performance level of the Search baselines. 
The RTU-PPO with simple memory variant achieved slightly higher average reward than the other PPO variants, but the gain is modest relative to the difference between the Search baselines. These results are not surprising because this task is much harder than the 2-biome variants. The increased rate of reward non-stationarity requires the agent to adapt quickly between tasks while also constructing state. The larger world size and limited FOV also requires the agent to explore and perform state construction to enable localization while remembering the cue. Contrasting our results with those of \cite{simmons2025deep} in  ForageWorld, we see the additional challenge posed by Forager's synchronous learning (i.e., a single stream of experience)---our CRL algorithms are still fairly data inefficient. 

This challenging variant of Forager opens many avenues for algorithmic development. The presence of partial observability and rapidly fluctuating rewards poses a unique challenge for continual learning with recurrent RL algorithms. The environment has a biome structure that makes it interesting for investigating temporal abstraction, option discovery, and model learning. Lastly, the larger world size highlights the need for effective continual exploration methods that enable agents to quickly navigate and discover the best biome. We could also make the environment more interesting and possibly more challenging by replacing the time-series behind each reward signal with timeseries from the real world, like the temperature in different cities.
This computationally inexpensive environment offers a unique challenge problem for driving further research progress in CRL.

\section{Conclusion}
In this paper we introduced a new continual learning testbed called Forager, that is ultra-lightweight and allows for rapid prototyping. This testbed is a family of environments, with configurable parameters on the size of the environment, level of partial observability and configurable objects with optionally non-stationary rewards. We proposed a variety of different tasks, provided sensible baselines for each that makes performance interpretable and tested several RL algorithms in these environments. We showed how algorithms shift from performing well to failing as the FOV is changed. We demonstrated that Forager can be used to study loss of plasticity and algorithms to mitigate it. We also showed how state construction via recurrent architectures play an important role in CRL. Finally, we showed that a surprising level of complexity can be introduced into this relatively bounded environment, simply through structured spawning and incorporating non-stationary observations and rewards. This environment is one where we can design agents to solve particular tasks, without the risk of overfitting, because new tasks are easy to create and likely to pose new challenges. Our experiments in Forager have highlighted that our existing algorithms have clear limitations, and that Forager can be both a challenge problem and a scientific testbed.

\subsubsection*{Broader Impact Statement}
\label{sec:broaderImpact}
This paper proposes a testbed for CRL research, that is neither realistic, nor of commercial interest. This work very much targets fundamental advances in RL algorithms and empirical practices, thus there are no concerns with broader impact.

\bibliography{main}
\bibliographystyle{rlj}

\raggedbottom

\beginSupplementaryMaterials

\section{Benchmarking compute and memory in Jelly Bean World and Forager}
\label{sec:benchmarking}

We benchmark Forager CPU implementation, Forager JAX implementation (Foragax), and Jelly Bean World in the large-scale foraging setup described in Section \ref{jbw-like} with a constant policy that always uses the up action for 10 million time steps. The metrics are averaged over 5 independent sequential runs on a desktop machine.

There are many factors such as world size, the complexity of environment dynamics, and hardware that impact benchmarking results. Consequently, these figures should be viewed as a relative performance comparison within the context of the large-scale foraging setup. 

\begin{table}[htbp]
\centering
\begin{tabular}{lccc}
\textbf{Environment} & \textbf{Wall time (h)} & \textbf{Speed (FPS)} & \textbf{Memory (GB)} \\
\hline
Forager & 0.02 & 159{,}879 & 0.1 \\
Foragax & 1.04 & 2{,}678 & 0.6 \\
Jelly Bean World & 5.70 & 487 & 13.2 \\
\end{tabular}
\caption{Resource utilization comparison across Forager, Foragax, and Jelly Bean World.}
\label{tab:resource_usage}
\end{table}

The benchmarking results in Table \ref{tab:resource_usage} demonstrate that for this large-scale foraging setup, Foragax was 5 times faster than JBW and Forager was 328 times faster than JBW. In Figure \ref{fig:jbw_grow}, we note that Forager and Foragax exhibited constant memory usage, while Jelly Bean World's memory usage increased linearly with the number of time steps.

\section{JBW-like experiments in Forager}
\label{jbw-like}

JBW introduced the idea of running experiments in 
an unbounded world to explore never-ending learning problems. However, it is the unbounded nature of JBW that also causes issues with unbounded  memory usage as the agent explores the world. A natural question to ask is whether large scale foraging experiments typically done in JBW produce similar learning dynamics in Forager with a fixed world size. To address this question, we set up a nearly-identical experiment in JBW and Forager to demonstrate that JBW-like large-scale foraging experiments can also be run in Forager. 

First, let us describe the setup of the experiment in Forager. We take inspiration from the non-episodic case study introduced by the original work~\citep{platanios2020jelly} and set up an experiment with rewarding objects (jelly beans) %
 and negative reward objects (onions),
with rewards of +1 and -1 respectively. The
objects spawn uniformly with 0.1 probability. The Forager world size is $1000 \times 1000$ with wrapping and objects respawn in random locations to approximate the infinite procedurally generated world in JBW.  This environment, which we denote as Forager Extra Large, is visualized in Figure \ref{fig:forager_extra_large_env}. The discount factor is 0.99.  We evaluate agents in our experiments using the exponential moving average of reward, with a decay of 0.999. This measure is similar to the reward rate used in previous work~\citep{platanios2020jelly} and measures an agent’s ability to adapt to the current circumstances.
The FOV is set to 11, resulting in $(11 \times 11 \times 2)$ observations with each channel encoding the absence or presence of an object. 

In both JBW and Forager we ran the same set of agents.
We ran DQN \citep{mnih2015human} with a convolutional layer with 16 kernels with size $3 \times 3$ with stride 1, followed by a hidden layer with width 64 and ReLU activation, then a final linear layer of size 4. We sweep the step size of each agent from $\{10^{-3}, 3\times 10^{-4}, 10^{-4}, 3\times 10^{-5}, 10^{-5}\}$ for 100 K steps averaging the total reward (area under the curve, AUC) over five runs (so called 1\% tuning~\citep{mesbahi2025position}). The hyperparameters are detailed in Tables \ref{tab:dqn_hypers_large_scale_foraging} and \ref{tab:dqn_select_hypers_large_scale_foraging}. We then evaluate each agent for 10 M steps with 30 independent trials with 30 random seeds, different from those used in the hyperparameter sweep (a two stage procedure~\citep{patterson2023empirical}). We also ran two baselines: one that selects actions uniformly (Random) and one that uses breadth first search to navigate greedily towards the closest jelly bean while avoiding onions (Greedy).

\begin{figure}[htbp]
    \centering
    \begin{subfigure}[b]{0.4\textwidth}
        \centering
        \includegraphics[width=0.9\textwidth]{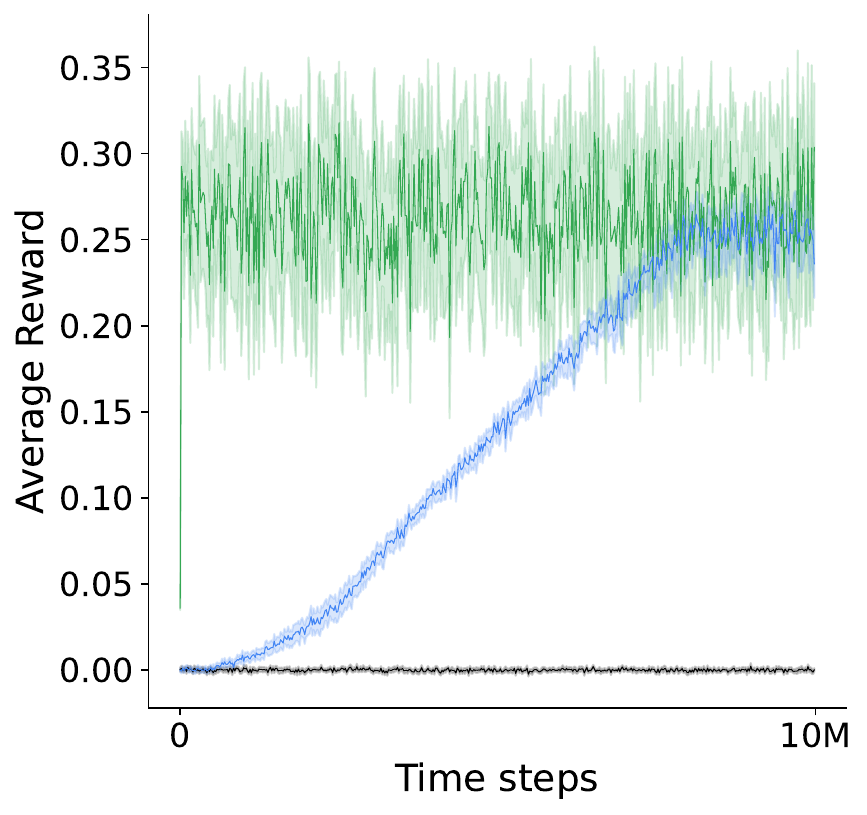}
        \caption{Jelly Bean World}
        \label{fig:jbw_large_scale_foraging}
    \end{subfigure}
    \hspace{0.5cm}
    \begin{subfigure}[b]{0.4\textwidth}
        \centering
        \includegraphics[width=0.9\textwidth]{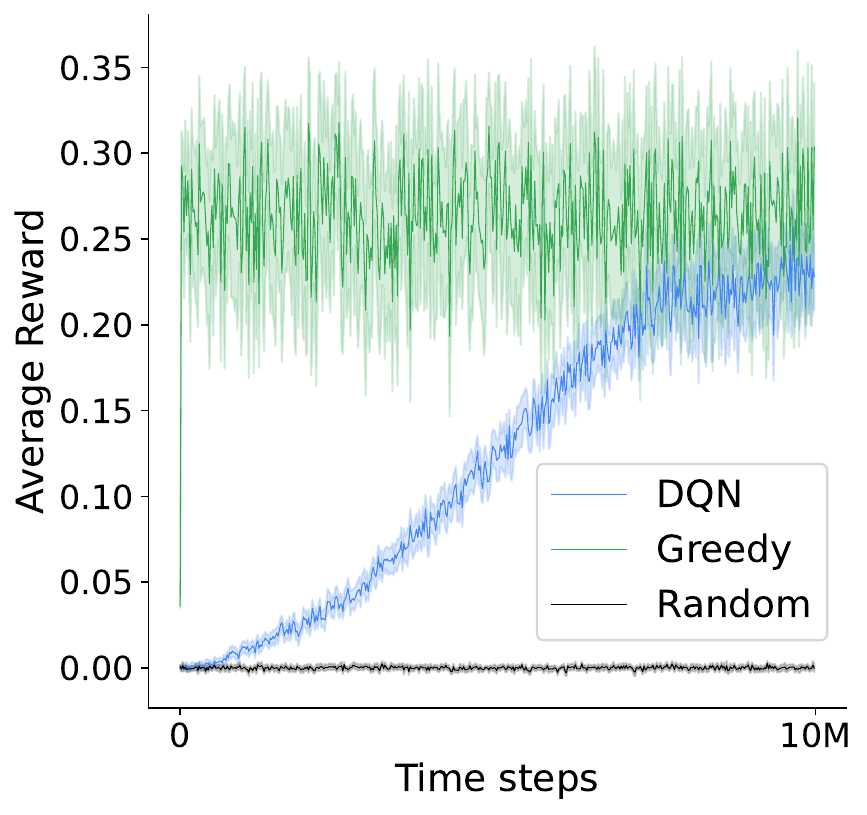}
        \caption{Forager Extra Large}
        \label{fig:forager_large_scale_foraging}
    \end{subfigure}
    \caption{Both JBW and Forager can be used to conduct large-scale gridworld foraging experiments. Results are averaged over 30 independent trials and the shaded regions are $95\%$ bootstrap CI.}
    \label{fig:jbw_forager_large_scale_foraging}
\end{figure}

Figure \ref{fig:jbw_forager_large_scale_foraging} shows that the learning curves of DQN and the two baselines were fairly similar between JBW and Forager. DQN eventually performed at a similar level to the Greedy search baseline in both environments. 
The environment and challenge posed by Forager appears to be quite similar to JBW. This is a large  Forager environment when you consider the size of the FOV.  
We also demonstrate that simple DQN agents are able to learn effective foraging for 10 M steps without underperforming later (no obvious loss of plasticity). The lightweight design of Forager allows us to run longer experiments and more independent trials with the same amount of compute. %
Overall, for large-scale foraging environments, Forager can serve as a more efficient drop-in replacement for JBW. 

\section{Experiment Details}
\label{sec:experiment_details}

\paragraph{Performance measure.} In the learning curves, the average reward is an unbiased exponentially weighted moving average of the per-step
reward with a decay of $0.999$, except in Figure~\ref{fig:forager_big_v5},
where the decay is $0.99999$.

\paragraph{Frequency units.} Frequencies in the hyperparameter tables are measured in environment steps, except for the target network update frequency, which is measured in gradient steps.

\paragraph{PPO.} PPO-based agents normalize advantages within each minibatch and clip the value loss.

\subsection{Large scale foraging experiment}
\label{sec:appendix_hyperparameters}    

\begin{table}[htbp]
    \caption{Hyperparameter choices and defaults for DQN in large scale foraging experiment}
    \begin{center}
        \begin{tabular}{lll}
            \multicolumn{1}{l}{\bf Hyperparameter}   &\multicolumn{1}{l}{\bf Choices}
            \\ \hline \\
            Step size                    & $\{10^{-3}, 3\times 10^{-4}, 10^{-4}, 3\times 10^{-5}, 10^{-5}\}$  \\  \\
            \multicolumn{1}{l}{\bf Hyperparameter} & \multicolumn{1}{l}{\bf Default} 
            \\ \hline \\
            Exploration Strategy &  $\epsilon$-greedy with linear decay \\
            $\epsilon$-greedy initial $\epsilon$                  & 1.0 \\
            $\epsilon$-greedy final $\epsilon$                    & 0.05 \\
            $\epsilon$-greedy decay percentage                    & 80\%  \\
            Update frequency              & 4 \\
            Minibatch size                     & 32 \\
            Replay memory size             & 10{,}000 \\
            Minimum replay history        & 32 \\
            Replay sampling strategy       & uniform \\
            Target network update frequency (gradient steps) & 128 \\
            Discount factor $\gamma$ & 0.99 \\
                                    Loss                            & Huber \\
            Optimizer & Adam \\
            Adam $\beta_1$ & 0.9 \\
            Adam $\beta_2$ & 0.999 \\
            Adam $\epsilon$ & $10^{-8}$ \\ 
        \end{tabular}
    \end{center}
    \label{tab:dqn_hypers_large_scale_foraging}
\end{table}

\begin{table}[htbp]
    \caption{Selected hyperparameters for DQN in large scale foraging experiment}
    \begin{center}
        \begin{tabular}{lll}
            \multicolumn{1}{l}{\bf Environment} &\multicolumn{1}{l}{\bf Forager Extra Large} &\multicolumn{1}{l}{\bf Jelly Bean World} 
            \\ \multicolumn{1}{l}{\bf Hyperparameter} &
            \\ \hline \\ 
            Step size          & $10^{-3} $ & $10^{-3}$          
        \end{tabular}
    \end{center}
    \label{tab:dqn_select_hypers_large_scale_foraging}
\end{table}

\vfill
\pagebreak

\subsection{Field of view experiment}
We compare against two breadth-first search baselines with full observability and avoid deathcaps: Search Oracle collects morels whenever they are available and oysters otherwise, while Search Nearest collects whichever is closest. We tuned hyperparameters using 2\% tuning with 5 independent trials, then evaluated each agent for 500{,}000 steps with 30 independent trials.

\begin{figure}[htbp]
    \centering
        \centering
        \includegraphics[width=\linewidth]{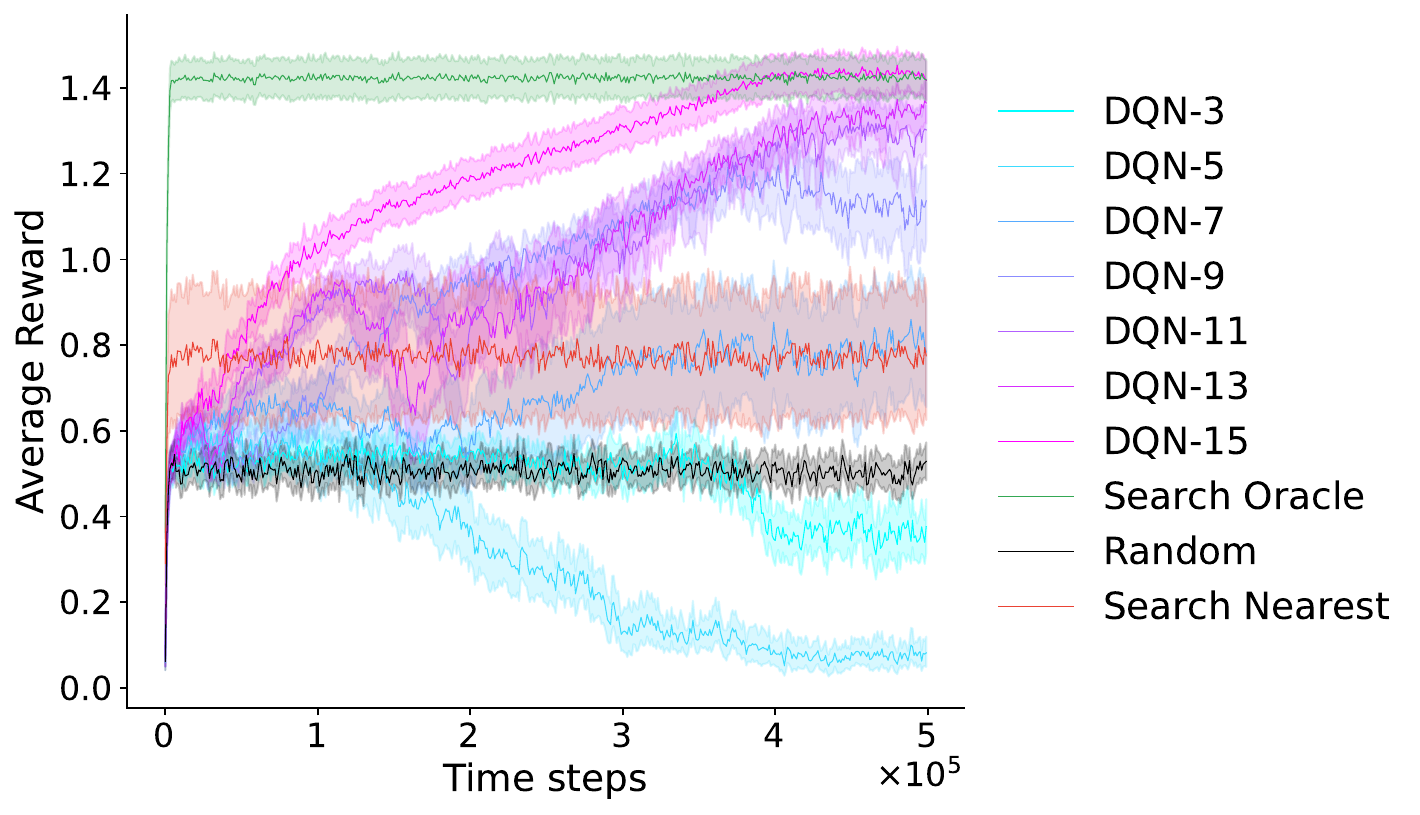}
    \caption{The effect of FOV on average reward learning curves of DQN agents. Agents with larger FOV sizes outperform those with smaller FOVs. Results are averaged over 30 independent trials and the shaded regions are $95\%$ bootstrap CI.}
    \label{fig:fov_learning_curves}
\end{figure}

\begin{table}[H]
    \caption{Hyperparameter choices and defaults for DQN in FOV experiments}
    \begin{center}
        \begin{tabular}{ll}
            \multicolumn{1}{l}{\bf Hyperparameter} & \multicolumn{1}{l}{\bf Choices}
            \\ \hline \\
            Step size & $\{10^{-3}, 3 \times 10^{-4}, 10^{-4}, 3 \times 10^{-5}, 10^{-5}\}$ \\
            Update frequency & $\{1, 4\}$ \\
            Target network update frequency (gradient steps) & $\{1, 128\}$ \\
            Adam $\beta_2$ & $\{0.9, 0.999\}$ \\
            Adam $\epsilon$ & $\{0.01, 10^{-8}\}$ \\  \\
            \multicolumn{1}{l}{\bf Hyperparameter} & \multicolumn{1}{l}{\bf Default} 
            \\ \hline \\
             Exploration Strategy &  $\epsilon$-greedy with linear decay \\
            $\epsilon$-greedy initial $\epsilon$                  & 1.0 \\
            $\epsilon$-greedy final $\epsilon$                    & 0.05 \\
            $\epsilon$-greedy decay percentage                    & 80\%  \\
            Minibatch size                     & 32 \\
            Replay memory size             & 10{,}000 \\
            Minimum replay history        & 32 \\
            Replay sampling strategy       & uniform \\
            Discount factor $\gamma$ & 0.99 \\
                                    Loss                            & Huber \\
            Optimizer & Adam \\
            Adam $\beta_1$ & 0.9 \\
        \end{tabular}
    \end{center}
    \label{tab:dqn_hypers_two_biome_fov_default}
\end{table}

\begin{table}[htbp]
    \caption{Selected hyperparameters for DQN in FOV experiment}
    \begin{center}
        \begin{tabular}{lllllllll}
            \multicolumn{1}{l}{\bf FOV}  & \multicolumn{1}{l}{\bf 3} & \multicolumn{1}{l}{\bf 5} & \multicolumn{1}{l}{\bf 7} & \multicolumn{1}{l}{\bf 9} & \multicolumn{1}{l}{\bf 11} & \multicolumn{1}{l}{\bf 13} & \multicolumn{1}{l}{\bf 15} & 
            \\ \multicolumn{1}{l}{\bf Hyperparameter} &&&&&&&
            \\ \hline \\ 
Step size & $10^{-3}$ & $10^{-3}$ & $10^{-4}$ & $3 \times 10^{-4}$ & $3 \times 10^{-4}$ & $3 \times 10^{-4}$ & $3 \times 10^{-4}$ \\
Update frequency & 4 & 4 & 1 & 4 & 4 & 4 & 4 \\
Target network \\ update frequency \\ (gradient steps) & 1 & 1 & 1 & 128 & 128 & 128 & 128 \\
Adam $\beta_2$ & 0.999 & 0.9 & 0.999 & 0.9 & 0.9 & 0.9 & 0.9 \\
Adam $\epsilon$ & 0.01 & 0.01 & 0.01 & $10^{-8}$ & $10^{-8}$ & $10^{-8}$ & $10^{-8}$ \\
        \end{tabular}
    \end{center}
    \label{tab:dqn_hypers_two_biome_fov_select}
\end{table}

\subsection{Never-ending relearning experiments}
\label{sec:neverending}
In the two-biome relearning environment, the reward mapping has a period of 500{,}000 steps, and switches every 250{,}000 steps. A collected mushroom respawns at a random location in the same biome after
 9--11 steps. Mushrooms that are not collected expire after 500 steps and respawn in the same way.
 
For DQN, the input observation is flattened and concatenated with the previous action and reward to form the input vector. If reward trace is enabled, an unbiased exponential moving average of the reward is also concatenated.

The DQN neural network architecture consists of a MLP with two hidden layers of 64 hidden units, each with LayerNorm and ReLU activation, followed by a linear projection to output 4 action-values. 

For DQN with ReDo, LayerNorm is applied after the ReLU activation, with dormancy scores computed on the LayerNorm outputs, and the previous action and reward are concatenated to the output of the first hidden layer rather than to the input.

The DRQN neural network adds a GRU with 64 hidden units between the two hidden layers. The input to the GRU is concatenated to its output to improve gradient flow.

The PPO neural network architecture consists of independent actor and critic networks with the same neural network body architecture. The neural network body has a single layer MLP encoder with 64 hidden units, LayerNorm, and Tanh activation which outputs latent features that are then concatenated with the previous action and reward. If reward trace is enabled, it is also concatenated. The concatenated embedding is followed by a hidden layer with 512 hidden units, and then a hidden layer with 64 hidden units with LayerNorm and Tanh activation. The actor has a linear projection head to output 4 action preferences. The critic has a linear projection head to output a value.

The RTU-PPO neural network follows the structure of the PPO neural network with one difference. The 512 unit hidden layer is replaced with Recurrent Trace Units (RTU) \citep{elelimy2024real} with 512 trace units, followed by ReLU. The input to the RTU is concatenated to its output to improve gradient flow. It is to be noted that RTU outputs two coupled values per trace unit, so the output directly from RTU is 1024.

\subsubsection{Hyperparameters} \label{sec:two_biome_hyper}

Tables \ref{tab:dqn_swept_hypers} and \ref{tab:dqn_defaults} contain the hyperparameters used for DQN variants; likewise, Tables \ref{tab:ppo_swept_hypers} and \ref{tab:ppo_defaults} contain the hyperparameters used for PPO variants.

The PT-DQN algorithm \cite{anand2023prediction} uses 64 hidden units for both the permanent and transient networks.

\begin{table}[H]
    \caption{Swept and agent-specific hyperparameters for DQN-based agents in never-ending relearning experiments. All agents sweep step size $\alpha \in \{3 \times 10^{-3}, 10^{-3}, 3 \times 10^{-4}, 10^{-4}, 3 \times 10^{-5}\}$ and $\epsilon \in \{0.05, 0.1, 0.25\}$ unless noted.}
    \begin{center}
        \begin{tabular}{lll}
            \multicolumn{1}{l}{\bf Agent} & \multicolumn{1}{l}{\bf Hyperparameter} & \multicolumn{1}{l}{\bf Choices / Selected}
            \\ \hline \\
            DQN                  & $\alpha$       & $3 \times 10^{-3}$ \\
                                 & $\epsilon$     & $0.1$ \\[4pt]
            DQN (CReLU)          & $\alpha$       & $3 \times 10^{-3}$ \\
                                 & $\epsilon$     & $0.1$ \\[4pt]
            DQN (L2)             & $\alpha$       & $3 \times 10^{-3}$ \\
                                 & $\epsilon$     & $0.1$ \\
                                 & $\lambda_{\text{L2}} \in \{10^{-2}, 10^{-3}, 10^{-4}, 10^{-5}\}$ & $10^{-5}$ \\[4pt]
            DQN (L2 Init)        & $\alpha$       & $10^{-3}$ \\
                                 & $\epsilon$     & $0.1$ \\
                                 & $\lambda_{\text{L2 Init}} \in \{10^{-2}, 10^{-3}, 10^{-4}, 10^{-5}\}$ & $10^{-5}$ \\[4pt]
              DQN (ReDo)           & $\alpha$       & $3 \times 10^{-3}$ \\
                                 & $\epsilon$     & $0.1$ \\
                                 & $\tau_{\text{ReDo}} \in \{0, 0.01, 0.025, 0.1\}$ & $0.01$ \\
                                 & ReDo frequency  & 10{,}000  \\[4pt]
            DQN (S\&P)           & $\alpha$       & $10^{-3}$ \\
                                 & $\epsilon$     & $0.25$ \\
                                 & Shrink factor $\in \{0.8, 0.9\}$   & $0.9$ \\
                                 & Noise scale $\in \{0.01, 0.001\}$ & $0.01$ \\
                                 & S\&P frequency  & 10{,}000  \\[4pt]
            DQN (Simple Memory)   & $\alpha$       & $3 \times 10^{-3}$ \\
                                 & $\epsilon$     & $0.1$ \\
                                 & Decay $\in \{0.9, 0.99\}$   & $0.9$ \\[4pt]
            DRQN                 & $\alpha$       & $10^{-4}$ \\
                                 & $\epsilon$     & $0.25$ \\
                                 & TBPTT length  & $16$ \\
                                 & Burn-in steps  & 16 \\[4pt]
            PT-DQN               & $\alpha$     & $10^{-3}$ \\
                                 & $\epsilon$                                              & $0.1$ \\
                                 & PT decay $\in \{0.55, 0.75, 0.95\}$                   & $0.95$ \\
                                 & PT $\alpha$ ratio $\in \{10^{-3}, 10^{-4}, 10^{-5}\}$ & $10^{-5}$ \\
                                 & PT frequency & 10{,}000 \\
                                 & PT replay memory size  & 10{,}000 \\
                                 & PT optimizer   & SGD \\
        \end{tabular}
    \end{center}
    \label{tab:dqn_swept_hypers}
\end{table}

\begin{table}[H]
    \caption{Shared default hyperparameters for all DQN-based agents in never-ending relearning experiments.}
    \begin{center}
        \begin{tabular}{ll}
            \multicolumn{1}{l}{\bf Hyperparameter} & \multicolumn{1}{l}{\bf Value}
            \\ \hline \\
            Exploration strategy            & $\epsilon$-greedy \\
            Update frequency                & 4 \\
            Minibatch size                  & 32 \\
            Replay memory size              & 1{,}000 \\
            Minimum replay history          & 50 \\
            Replay sampling strategy        & uniform \\
            Target network update frequency (gradient steps) & 128 \\
            Discount factor $\gamma$        & 0.99 \\
                        Loss                            & Huber \\
            Optimizer                       & Adam \\
            Adam $\beta_1$                  & 0.9 \\
            Adam $\beta_2$                  & 0.999 \\
            Adam $\epsilon$                 & $10^{-5}$ \\
        \end{tabular}
    \end{center}
    \label{tab:dqn_defaults}
\end{table}

\begin{table}[H]
    \caption{Swept and agent-specific hyperparameters for PPO-based agents in never-ending relearning experiments. All agents sweep actor step size $\alpha_a \in \{10^{-3}, 3 \times 10^{-4}, 10^{-4}\}$, Critic step size scale factor $\in \{0.1, 1.0, 10.0\}$ (the critic learning rate $\alpha_c$ is set to $\alpha_a \times \text{Critic step size scale factor}$), and entropy coefficient $\in \{0.01, 0.1, 1.0\}$.}
    \begin{center}
        \begin{tabular}{lll}
            \multicolumn{1}{l}{\bf Agent} & \multicolumn{1}{l}{\bf Hyperparameter} & \multicolumn{1}{l}{\bf Choices / Selected}
            \\ \hline \\
            PPO                  & $\alpha_a$      & $3 \times 10^{-4}$ \\
                                 & Critic step size scale factor & $0.1$ \\
                                 & Entropy coefficient   & $0.01$ \\[4pt]
            PPO (L2)             & $\alpha_a$      & $10^{-3}$ \\
                                 & Critic step size scale factor & $0.1$ \\
                                 & Entropy coefficient   & $0.01$ \\
                                 & $\lambda_{\text{L2}} \in \{10^{-2}, 10^{-3}, 10^{-4}\}$ & $10^{-4}$ \\[4pt]
            PPO (L2 Init)        & $\alpha_a$      & $10^{-3}$ \\
                                 & Critic step size scale factor & $0.1$ \\
                                 & Entropy coefficient   & $0.01$ \\
                                 & $\lambda_{\text{L2 Init}} \in \{10^{-2}, 10^{-3}, 10^{-4}\}$ & $10^{-4}$ \\[4pt]
            PPO (S\&P)           & $\alpha_a$      & $10^{-3}$ \\
                                 & Critic step size scale factor & $1.0$ \\
                                 & Entropy coefficient   & $0.01$ \\
                                 & Shrink factor $\in \{0.8, 0.9\}$   & $0.9$ \\
                                 & Noise scale $\in \{0.01, 0.001\}$ & $0.01$ \\
                                 & S\&P frequency   & 10{,}000 \\[4pt]
            PPO (Simple Memory)   & $\alpha_a$      & $10^{-3}$ \\
                                 & Critic step size scale factor & $0.1$ \\
                                 & Entropy coefficient   & $0.01$ \\
                                 & Decay $\in \{0.9, 0.99\}$   & $0.9$ \\[4pt]
            RTU-PPO        & $\alpha_a$      & $3 \times 10^{-4}$ \\
                                 & Critic step size scale factor & $10.0$ \\
                                 & Entropy coefficient   & $0.1$ \\
        \end{tabular}
    \end{center}
    \label{tab:ppo_swept_hypers}
\end{table}

\begin{table}[H]
    \caption{Shared default hyperparameters for all PPO-based agents in never-ending relearning experiments.}
    \begin{center}
        \begin{tabular}{ll}
            \multicolumn{1}{l}{\bf Hyperparameter} & \multicolumn{1}{l}{\bf Value}
            \\ \hline \\
            Rollout steps                & 2{,}048 \\
            Epochs                       & 4 \\
            Number of mini-batches       & 32 \\
            Clipping $\epsilon$          & 0.2 \\
            Max gradient norm            & 0.5 \\
            Value function coefficient   & 0.5 \\
            GAE $\lambda$                & 0.95 \\
            Discount factor $\gamma$     & 0.99 \\
            Optimizer                    & Adam \\
            Adam $\beta_1$               & 0.9 \\
            Adam $\beta_2$               & 0.999 \\
            Adam $\epsilon$              & $10^{-5}$ \\
        \end{tabular}
    \end{center}
    \label{tab:ppo_defaults}
\end{table}

\subsection{Unending tasks experiment}
\label{sec:unending}

The reward function for each biome used is $r(t) = \sum_{n=1}^N \left(a_n \cos\left(\frac{2\pi n}{T}\left\lfloor\frac{t}{w}\right\rfloor\right)  + b_n \sin\left(\frac{2\pi n}{T}\left\lfloor\frac{t}{w}\right\rfloor\right)\right)$, with parameters $N = 10$, reward repeat $w = 1000$ (the reward is held constant for 1000 environment steps). Each biome independently samples the parameters $a_n, b_n \sim \mathcal{N}\left(0, \frac{1}{n^2}\right)$, the period $T$ is sampled uniformly from $\{10, 11, \ldots, 1000\}$ (the reward function cycles after $Tw$ environment steps). Each biome's raw Fourier sum is then min-max normalized to $[-1, 1]$ using the extrema over one full cycle, so every mushroom species has the same reward range regardless of the sampled $a_n, b_n$; the sampled parameters controls the shape of the reward curve rather than its amplitude. All the reward time series in the environment are centered at each time step by subtracting the mean reward over all mushrooms to ensure that there is always at least one mushroom species with non-negative rewards. Mushrooms respawn randomly within a biome with respawn time $m$ sampled uniformly from $\{9, 10\}$ steps. Walls are scattered through the environment to aid an agent to locate itself within the large world and force it to learn how to navigate around them. 

We use 10 independent trials with 10\%-tuning to identify the best hyperparameters. We evaluate each agent for 10 M steps with 30 independent trials.

The PPO and RTU-PPO neural network architectures are modified by the addition of a convolutional vision encoder to process RGB images. The convolutional vision encoder consists of a 
pointwise convolutional layer (16 filters with kernel size $1 \times 1$ with a stride of 1, followed by LayerNorm and Tanh activation), a
convolutional layer (16 filters with kernel size $1 \times 3$ with a stride of 1, followed by LayerNorm and Tanh activation). In preliminary experiments, a pointwise convolutional layer provided useful inductive bias for learning. The resulting vision features are flattened into a 1D vector. Following the vision encoder is a linear layer with 64 hidden units, LayerNorm, and Tanh activation. Then the global cue vector is concatenated along with the previous action and reward to the output of the encoder. For PPO, this is followed by a linear layer with 1024 hidden units with LayerNorm and Tanh activation. For RTU-PPO, this is followed by a layer of Recurrent Trace Units (RTU) \citep{elelimy2024real} with 1024 trace units, followed by ReLU activations. The input to the RTU is concatenated to its output to improve gradient flow. In both network architectures, this is followed by another linear layer with 64 hidden units with LayerNorm and Tanh activation. The actor has a linear projection head to output 4 action preferences. The critic has a linear projection head to output a value.

\begin{table}[H]
    \caption{Hyperparameter choices and defaults for PPO in Unending Tasks Forager experiment}
    \begin{center}
        \begin{tabular}{llll}
            \multicolumn{1}{l}{\bf Hyperparameter} & \multicolumn{1}{l}{\bf Choices} & \multicolumn{1}{l}{\bf Selected} & \multicolumn{1}{l}{\bf Selected (Simple Memory)} 
            \\ \hline \\
            Actor step size & $\{10^{-3}, 3 \times 10^{-4}, 10^{-4}\}$ & $10^{-4}$ & $10^{-4}$ \\
            Entropy coefficient & $\{0.01, 0.1, 1.0\}$ & $0.1$ & $0.1$ \\
            Critic step size scale factor & $\{0.1, 1.0, 10\}$ & $0.1$  & $1$ \\
            Decay & $\{0.9, 0.99\}$  & N/A & $0.99$\\ \\
            \multicolumn{1}{l}{\bf Hyperparameter} & \multicolumn{1}{l}{\bf Default}
            \\ \hline \\
            Rollout steps                & 128 \\
            Epochs per update            & 4 \\
            Number of minibatches        & 1 \\
            Clipping $\epsilon$          & 0.2 \\
            Max gradient norm            & 0.5 \\
            Value function coefficient   & 0.5 \\
            GAE $\lambda$                & 0.95 \\
            Discount factor $\gamma$     & 0.99 \\
            Optimizer                    & Adam \\
            Adam $\beta_1$               & 0.9 \\
            Adam $\beta_2$               & 0.999 \\
            Adam $\epsilon$              & $10^{-5}$ \\
        \end{tabular}
    \end{center}
    \label{tab:ppo_hypers_two_biome_fov_default}
\end{table}

\begin{table}[H]
    \caption{Hyperparameter choices and defaults for RTU-PPO in Unending Tasks Forager experiment}
    \begin{center}
        \begin{tabular}{llll}
            \multicolumn{1}{l}{\bf Hyperparameter} & \multicolumn{1}{l}{\bf Choices} & \multicolumn{1}{l}{\bf Selected} & \multicolumn{1}{l}{\bf Selected (Simple Memory)}
            \\ \hline \\
            Actor step size & $\{10^{-3}, 3 \times 10^{-4}, 10^{-4}\}$ & $10^{-4}$ & $10^{-4}$ \\
            Entropy coefficient & $\{0.01, 0.1, 1.0\}$ & $0.1$ & $0.1$ \\
            Critic step size scale factor & $\{0.1, 1.0, 10\}$ & $1$ & $1$ \\
            Decay & $\{0.9, 0.99\}$  & N/A & $0.9$\\ \\

            \multicolumn{1}{l}{\bf Hyperparameter} & \multicolumn{1}{l}{\bf Default}
            \\ \hline \\
            Rollout steps                & 128 \\
            Epochs per update            & 4 \\
            Number of minibatches        & 1 \\
            Clipping $\epsilon$          & 0.2 \\
            Max gradient norm            & 0.5 \\
            Value function coefficient   & 0.5 \\
            GAE $\lambda$                & 0.95 \\
            Discount factor $\gamma$     & 0.99 \\
            Optimizer                    & Adam \\
            Adam $\beta_1$               & 0.9 \\
            Adam $\beta_2$               & 0.999 \\
            Adam $\epsilon$              & $10^{-5}$ \\
        \end{tabular}
    \end{center}
    \label{tab:ppo_rtu_hypers_two_biome_fov_default}
\end{table}

\end{document}